\documentclass[journal]{IEEEtran}

\usepackage{amsmath,amssymb,amsfonts}
\usepackage{graphicx}
\usepackage{cite}
\usepackage{booktabs}
\usepackage{multirow}
\usepackage{enumitem}
\usepackage{url}
\usepackage{ulem}
\usepackage[hidelinks,hypertexnames=false,colorlinks=true,linkcolor=blue!80!black,citecolor=red!70!black,urlcolor=black]{hyperref}
\usepackage{orcidlink}

\newcommand{\settablefont}{\fontsize{7}{10}\selectfont}
\newcommand{\eg}{\textit{e.g.}}
\newcommand{\ie}{\textit{i.e.}}

\begin{document}

    \title{PanoBench: A Panoptic Occupancy Prediction Benchmark for Autonomous Driving}

    \author{
        Yi~Feng\textsuperscript{*}\,\orcidlink{0009-0005-4885-0850},
        Zizhan~Guo\textsuperscript{*}\,\orcidlink{0009-0003-7360-2192},
        Junwu~E\,\orcidlink{0009-0000-1288-2360},
        Yu~Ma\,\orcidlink{0009-0000-8536-5444},
        Mengtan~Zhang\,\orcidlink{0009-0003-3468-7680},
        Jin~Wu\,\orcidlink{0000-0001-5930-4170},~\IEEEmembership{Member,~IEEE},
        Wei~Ye\,\orcidlink{0000-0002-3784-7788},~\IEEEmembership{Member,~IEEE},
        Hanli~Wang\,\orcidlink{0000-0002-9999-4871},~\IEEEmembership{Senior Member,~IEEE},
        Qijun~Chen\,\orcidlink{0000-0001-5644-1188},~\IEEEmembership{Senior Member,~IEEE},
        Danwei~Wang\,\orcidlink{0000-0003-3400-0079},~\IEEEmembership{Life Fellow,~IEEE},
        and Rui~Fan\textsuperscript{\textdagger}\,\orcidlink{0000-0003-2593-6596},~\IEEEmembership{Senior Member,~IEEE}
        \thanks{\textsuperscript{*}Equal contribution.}
        \thanks{\textsuperscript{\textdagger}Corresponding author.}
    }

\maketitle

\begin{abstract}
            Panoptic occupancy provides a unified representation of 3D scene geometry, semantic categories, and object identities, but its development is constrained by the scarcity of large-scale instance-aware occupancy annotations.
            Existing benchmarks predominantly provide semantic occupancy labels reconstructed from incomplete sensor observations, while available panoptic ones are limited in domain coverage or annotation completeness.
            This article introduces PanoBench, a camera-centric panoptic occupancy prediction benchmark with multi-resolution, high-fidelity, and temporally consistent panoptic annotations.
            PanoBench unifies diverse data sources from multiple real-world datasets (\eg, nuScenes, KITTI-360)  and virtual driving simulators (\eg, CARLA), providing canonical data interfaces and unified evaluation protocols.
            For the synthetic branch, PanoBench-CarlaOcc derives complete panoptic geometry directly from simulated meshes and scene layouts.
            For the real-world branch, a new panoptic occupancy generation pipeline is proposed to construct PanoBench-nuScenes and PanoBench-KITTI360.
            The pipeline reconstructs background point clouds with geometric priors and topology constraints, generates foreground mesh assets from multi-frame multi-view images, and fuses both components into a sparse voxel-based panoptic representation.
            To ensure reasonable evaluation of existing models, a unified valid-mask protocol is developed to evaluate complete thing instances once they have been observed, while scoring stuff regions conservatively at the voxel level.
            Furthermore, extensive analyses of observation fidelity, geometric completeness, spatial continuity, and temporal consistency demonstrate the superiority of PanoBench over existing public occupancy prediction benchmarks.
            Finally, this study provides fair and reproducible comparisons of representative baselines on panoptic occupancy prediction, semantic occupancy prediction, and 4D occupancy forecasting, facilitating further advancements in autonomous driving perception systems.
            The proposed benchmark is available at \url{https://mias.group/PanoBench}.

\end{abstract}


\IEEEpeerreviewmaketitle

    \begin{figure*}[!t]
        \centering
        \includegraphics[width=0.99\textwidth]{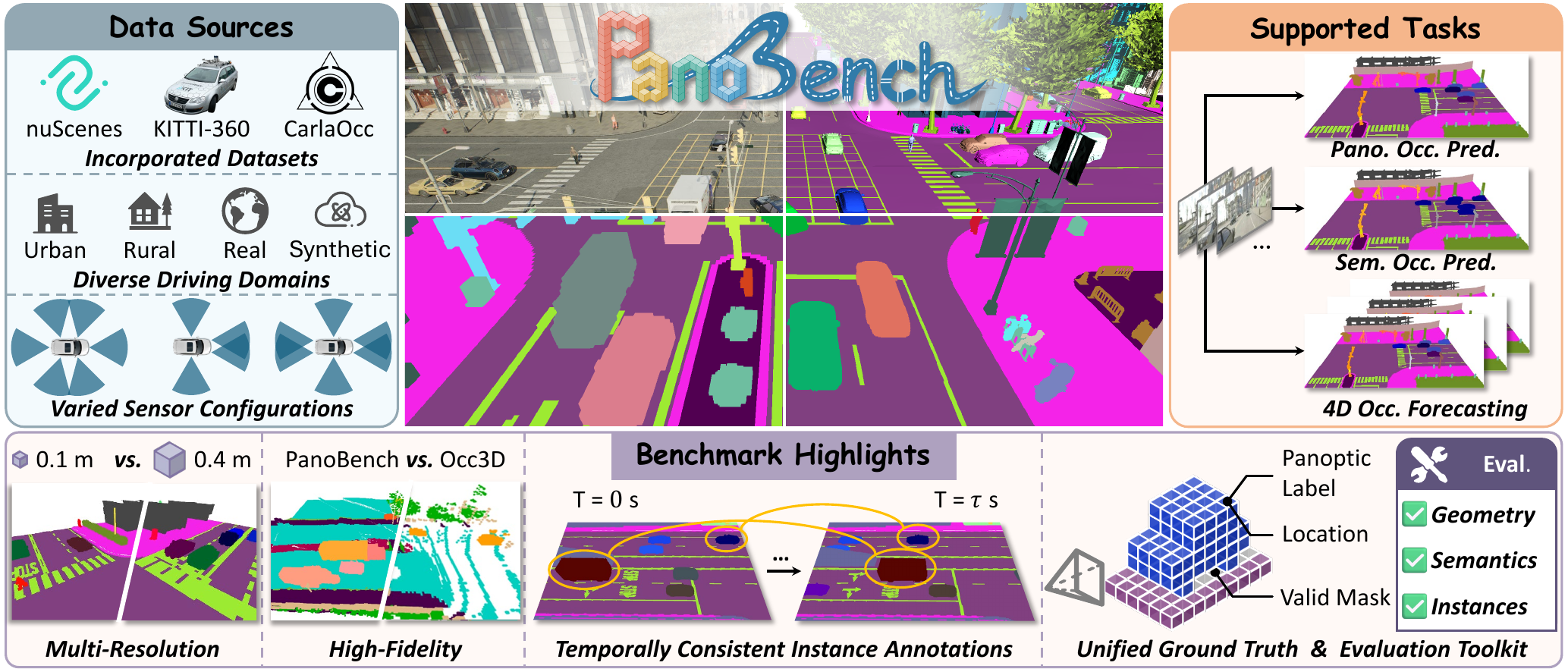}
        \caption{Overview of PanoBench, a unified benchmark for camera-centric panoptic occupancy prediction.
        By incorporating diverse driving datasets from both synthetic and real-world domains, PanoBench provides multi-resolution, high-fidelity, and temporally consistent instance annotations, establishing a unified testbed for semantic occupancy prediction, panoptic occupancy prediction, and 4D occupancy forecasting.}
        \label{fig.teaser}
    \end{figure*}

    \section{Introduction}
    \label{sec:introduction}

    3D scene understanding is a fundamental capability for autonomous driving systems, as it supports spatial reasoning about free space, occupied regions, semantic categories, and surrounding traffic participants~\cite{bevformer2022li,sima2023occnet,panoocc2024wang}.
    Compared with object-centric perception tasks such as 3D object detection and tracking~\cite{centerpoint2021yin,detr3d2021wang}, occupancy prediction provides a more complete description of the driving scene by representing the environment as a structured 3D field~\cite{occ3d2023tian,openoccupancy2023wang}.
    Recent semantic occupancy prediction benchmarks and methods~\cite{semantickitti2019behley,tpvformer2023huang,occ3d2023tian,surroundocc2023wei,openoccupancy2023wang,sscbench2024li,hisop2026li,reo2026zhuang} have substantially advanced this field by requiring models to infer voxel-level geometry and semantic labels from camera or LiDAR observations.
    These efforts have made occupancy prediction an important formulation for vision-centric autonomous driving perception.

    However, semantic occupancy alone is insufficient for holistic 3D scene understanding~\cite{pasco2024cao,panoocc2024wang,sparseocc2024liu,flashocc2024yu}.
    It predicts voxel occupancy and semantic category, but does not distinguish different object instances within the same semantic class~\cite{panoptic2019kirillov,pasco2024cao,panoocc2024wang}.
    As a result, semantic occupancy prediction benchmarks cannot directly evaluate whether a model separates adjacent vehicles, preserves complete object shapes, or maintains instance-level consistency across temporal observations~\cite{pasco2024cao,sparseocc2024liu,flashocc2024yu,carlaocc2026feng}.
    This limitation becomes critical for camera-centric perception, where scene elements might be partially observed and must be inferred from multi-view and temporal cues~\cite{bevformer2022li,cam4docc2024ma}.
    Panoptic occupancy prediction bridges this gap by assigning each occupied voxel both a semantic category and an instance identity, thereby extending panoptic segmentation from image regions to dense 3D space~\cite{panoptic2019kirillov,pasco2024cao,panoocc2024wang,sparseocc2024liu,flashocc2024yu}.

    The main bottleneck for panoptic occupancy prediction lies in benchmark construction.
    Existing real-world occupancy datasets~\cite{semantickitti2019behley,occ3d2023tian,surroundocc2023wei,openoccupancy2023wang,sscbench2024li,uniscenev22026li} are mainly designed for semantic occupancy prediction, and their annotations are generally reconstructed from sparse, noisy, and visibility-limited sensor observations.
    Such annotations are difficult to extend to reliable instance-level 3D supervision, especially for complete foreground objects and large background regions under occlusions.
    Synthetic environments can provide more complete geometry through mesh-level scene access~\cite{carla2017dosovitskiy,carlaocc2026feng}, but a synthetic-only benchmark cannot fully characterize the appearance diversity, sensor characteristics, and annotation challenges of real-world autonomous driving data.
    Therefore, a comprehensive benchmark should combine high-fidelity synthetic annotations with reconstructed real-world panoptic occupancy annotations under a unified task definition and evaluation protocol.

    This article extends our previous work~\cite{carlaocc2026feng} from a synthetic panoptic occupancy benchmark to a unified synthetic and real-world benchmark for camera-centric panoptic occupancy prediction.
    Our previous study introduced CarlaOcc, a large-scale synthetic dataset with high-fidelity panoptic occupancy annotations generated from simulated meshes.
    This article systematizes it as PanoBench-CarlaOcc and further constructs two real-world panoptic occupancy datasets, PanoBench-nuScenes and PanoBench-KITTI360, based on nuScenes~\cite{nuscenes2020caesar} and KITTI-360~\cite{kitti3602023liao}.
    Together, these datasets form PanoBench, a comprehensive benchmark covering both controlled synthetic scenes and complex real-world driving scenarios.
    As shown in Fig.~\ref{fig.teaser}, the unified benchmark provides multi-resolution, high-fidelity occupancy labels, temporally consistent instance annotations, and a unified ground-truth format for fair evaluation of 3D/4D semantic and panoptic occupancy prediction.

    PanoBench is organized into a synthetic branch and a real-world branch.
    For the synthetic branch, CarlaOcc is reorganized as PanoBench-CarlaOcc under the PanoBench protocol.
    Its ground truth is generated directly from simulated meshes, providing complete, physically consistent panoptic geometry without relying on sparse sensor accumulation.
    We standardize its spatial ranges, sparse storage, panoptic identity encoding, and valid masks so that the same model interfaces and evaluation tools apply to both synthetic and real-world~data.

    For the real-world branch, PanoBench annotates panoptic occupancy by separating static geometry from foreground instances before integrating them into a common voxel space.
    Background reconstruction combines temporally aggregated semantic LiDAR observations, HD map-guided ground surface lifting, topology-aware flat region recovery, and geometry-preserving filtering of non-planar structures.
    Foreground reconstruction mines high-quality, viewpoint-diverse instance observations from synchronized multi-view image sequences and converts them into reusable complete meshes.
    These meshes are aligned with annotated 3D boxes and fused with the reconstructed background, yielding sparse panoptic voxels in which semantic categories and instance identities are stored.
    By decoupling static background recovery from foreground instance completion, this design reduces reliance on direct sensor accumulation for dynamic objects and provides more effective supervision for occluded or partially observed traffic participants.

    Beyond dataset construction, PanoBench establishes a unified camera-centric evaluation protocol.
    It uses temporally aware valid masks to retain complete thing instances supported by historical observations while conservatively evaluating stuff geometry within the observable camera space.
    This protocol makes the benchmark sensitive to whether a model can exploit temporal evidence to preserve object completeness and identity, rather than only recovering surfaces visible at the target frame.
    As a result, PanoBench supports a more comprehensive study of instance separation, shape completion, occlusion-aware reconstruction, temporal consistency, and domain robustness.

    The main contributions are summarized as follows:
    \begin{itemize}[leftmargin=*]
    \item We present \textit{\textbf{PanoBench}}, a comprehensive camera-centric panoptic occupancy benchmark that combines two curated real-world datasets, PanoBench-nuScenes and PanoBench-KITTI360, with the high-fidelity synthetic PanoBench-CarlaOcc dataset, enabling unified study of annotation fidelity, real-world reconstruction, and sim-to-real transfer.
    \item We establish a unified panoptic occupancy construction framework that systematizes CarlaOcc and introduces real-world background reconstruction, foreground instance generation, and panoptic fusion under a canonical annotation format, supporting complete instance-level supervision for partially observed scenes.
    \item We provide a standardized camera-centric evaluation protocol for complete panoptic occupancy, rather than for surfaces that are only visible in the current frame. The protocol comprises sparse panoptic labels and temporally aware valid masks, under which we benchmark representative methods for semantic occupancy prediction, panoptic occupancy prediction, and 4D occupancy forecasting.
    \end{itemize}

    The remainder of this article is organized as follows:
    Sect.~\ref{sec:related_work} reviews semantic occupancy prediction benchmarks, vision-based occupancy prediction, and panoptic occupancy prediction.
    Sect.~\ref{sec:method} defines the panoptic occupancy prediction task and describes the construction of PanoBench, including synthetic data generation, real-world reconstruction, and valid mask generation.
    Sect.~\ref{sec:experiments} presents the experimental setup, occupancy annotation quality analyses, ablations of the construction pipeline, occupancy prediction benchmarks, and extended studies on voxel resolution, sim-to-real transfer, and storage efficiency.
    Finally, Sect.~\ref{sec:conclusion} concludes the article and discusses future work.

    \section{Related Work}
    \label{sec:related_work}

    \subsection{Semantic Occupancy Prediction Benchmarks}
    The semantic occupancy prediction task aims to recover occupied geometry and voxel-wise semantics from incomplete sensor observations.
    SSCNet~\cite{sscnet2017song} pioneers end-to-end semantic scene completion by predicting dense voxel occupancy and semantics from a single depth observation.
    SemanticKITTI~\cite{semantickitti2019behley} establishes an early outdoor benchmark by registering sequential LiDAR scans into a common coordinate frame.
    However, its limited scene diversity and forward-facing setup make it difficult to assess the surround-view, camera-centric perception required by contemporary autonomous driving systems.
    MonoScene~\cite{monoscene2022cao} introduces monocular camera-based semantic scene completion, motivating subsequent vision-centric occupancy methods.
    Later benchmarks, such as SurroundOcc~\cite{surroundocc2023wei}, Occ3D~\cite{occ3d2023tian}, and OpenOccupancy~\cite{openoccupancy2023wang}, extend occupancy supervision from forward scene completion to surrounding perception.
    By combining techniques such as Poisson surface reconstruction~\cite{kazhdan2006poisson}, occlusion reasoning, and manual label refinement, these datasets provide denser and more detailed occupancy annotations for complex driving scenes.
    SSCBench~\cite{sscbench2024li} further aligns labels and evaluation settings across various real-world datasets such as KITTI-360~\cite{kitti3602023liao}, nuScenes~\cite{nuscenes2020caesar}, and Waymo~\cite{waymo2020sun}, enabling controllable comparisons across sensor configurations, geographic regions, and data domains.

    Despite these advances, real-world occupancy annotations inevitably suffer from incomplete geometries due to the sparsity of LiDAR point clouds and pervasive occlusions.
    To mitigate this limitation, several works~\cite{collaborative2024song, motionsc2022wilson} leverage virtual engines to generate high-quality simulated data.
    Our prior work, CarlaOcc~\cite{carlaocc2026feng}, uses Unreal Engine 5~\cite{unrealengine5} to construct occupancy ground truth directly from virtual mesh assets, yielding high-fidelity, resolution-agnostic, and physically consistent scene representations.
    Nevertheless, simulation-based benchmarks still face a substantial domain gap from real-world driving environments.
    PanoBench addresses this limitation by bridging high-quality simulation-based occupancy annotations with more realistic and diverse driving scenarios, providing a more reliable benchmark for evaluating semantic occupancy prediction under practical domain shifts.

    \subsection{Vision-Based Occupancy Prediction}
    Vision-based occupancy prediction estimates dense 3D geometry and semantics from camera observations.
    A common line of work lifts image features into structured 3D representations.
    LSS~\cite{lss2020philion} establishes the forward-projection paradigm that lifts multi-view image features and splats them into a bird's-eye-view (BEV) grid, and subsequent detectors such as BEVDepth~\cite{bevdepth2023li} further improve depth-aware view transformation.
    TPVFormer~\cite{tpvformer2023huang} augments the BEV representation by factorizing the volume into three orthogonal feature planes, while VoxFormer~\cite{voxformer2023Li} starts from sparse visible voxel proposals and densifies them for monocular semantic scene completion.
    OccFormer~\cite{occformer2023Zhang} decomposes heavy 3D processing into efficient Transformer pathways, while SurroundOcc~\cite{surroundocc2023wei} progressively upsamples multi-scale 3D volume features for multi-camera occupancy prediction.
    FB-OCC~\cite{fbocc2023li} couples forward and backward view transformations for denser 3D features, while FlashOcc~\cite{flashocc2023yu} replaces costly 3D convolutions with a channel-to-height plugin for efficient occupancy decoding.
    Another line of work reduces the cost of dense voxel prediction by exploiting scene sparsity and adaptive representations.
    GaussianFormer~\cite{gaussianformer2024huang} represents driving scenes with sparse semantic Gaussians, and GaussianFormer-2~\cite{gaussianformer2} improves this formulation with probabilistic Gaussian superposition.
    SparseOcc~\cite{sparseocc2024liu} predicts semantic and instance occupancy from a fully sparse 3D representation and introduces RayIoU to reduce the depth-axis ambiguity of voxel-level mIoU.
    OPUS~\cite{opus2024wang} further formulates occupancy as a sparse set prediction problem, directly predicting occupied locations and classes with learnable queries.
    These methods improve computational efficiency and the compactness of 3D occupancy representations.
    Beyond fully supervised training, complementary self-supervised approaches based on neural radiance fields~\cite{bts2023wimbauer,kyn2024li,selfocc2024huang,vipocc2025feng} further reduce reliance on dense occupancy annotations.
    In addition to current-frame prediction, 4D occupancy forecasting estimates future occupancy from past observations~\cite{worldmodel2025kong}.
    Cam4DOcc~\cite{cam4docc2024ma} establishes a camera-only forecasting benchmark, OccWorld~\cite{occworld2024zheng} and Drive-OccWorld~\cite{driveoccworld2025yang} learn occupancy world models for joint forecasting and planning, and SparseWorld~\cite{sparseworld2026dang} and SparseWorld-TC~\cite{sparseworldtc2025du} build sparse occupancy world models, the latter being conditioned on the ego trajectory.
    However, existing supervision and evaluation protocols remain dominated by semantic occupancy~\cite{occscene2025li,uniscenev22026li}, leaving panoptic instance quality and cross-dataset panoptic generalization underexplored.
    PanoBench is designed to support these comparisons through unified semantic and instance-aware annotations across synthetic and real-world domains.

    \subsection{Panoptic Occupancy Prediction}
    Panoptic segmentation unifies semantic segmentation and instance segmentation by assigning each pixel both a category and, for thing classes, an instance identity~\cite{panoptic2019kirillov}.
    Extending this idea to 3D scene understanding is important because autonomous driving systems must reason about both stuff regions and individual traffic participants.
    Early studies~\cite{semantickitti_panoptic2021behley,panoptic_polarnet2021zhou,efficientlps2021sirohi,fourdpls2021aygun} investigate LiDAR-based panoptic perception.
    PaSCo~\cite{pasco2024cao} introduces panoptic scene completion, extending semantic scene completion with instance-aware 3D prediction and uncertainty estimation.
    PanoOcc~\cite{panoocc2024wang} formulates camera-based 3D panoptic segmentation in a unified occupancy representation and aggregates multi-frame multi-view information through voxel queries.
    SparseOcc~\cite{sparseocc2024liu} and Panoptic-FlashOcc~\cite{flashocc2024yu} further show that panoptic occupancy can be approached through sparse queries or efficient instance-center prediction.
    Recent studies~\cite{trackocc2025chen,occtrack3602026lin} further investigate 4D panoptic occupancy tracking, covering both camera-based methods and surround-view fisheye benchmarks.
    These works establish important baselines for instance-aware occupancy prediction, but public panoptic occupancy datasets remain scarce.
    Our previous work CarlaOcc~\cite{carlaocc2026feng} addresses this issue in the synthetic domain by generating high-fidelity panoptic occupancy annotations from simulated meshes.
    However, a synthetic-only benchmark cannot fully reflect real-world sensor characteristics, annotation incompleteness, and domain shifts.
    PanoBench extends CarlaOcc into a comprehensive benchmark by systematizing it as PanoBench-CarlaOcc and adding PanoBench-nuScenes and PanoBench-KITTI360 with reconstructed panoptic occupancy annotations.
    This enables a more systematic study of panoptic occupancy prediction, including real-world reconstruction quality, cross-domain generalization, and camera-centric evaluation.

    \section{Methodology}
    \label{sec:method}

    \subsection{Task Formulation}
    \label{subsec:task_formulation}
    Let $\mathcal{I}_t=\{I_{\tau}^{k}\}_{\tau=t-T+1:t}^{k=1:K}$ be the input observations of multi-view RGB image sequences, where $K$ is the number of cameras and $T$ is the number of historical frames used for prediction.
    The camera intrinsics, extrinsics, and ego poses are collectively denoted by $\mathcal{G}_t$.
    The goal of panoptic occupancy prediction is to estimate a 3D panoptic scene representation in an ego-centric voxel space:
    \begin{equation}
        f(\mathcal{I}_t,\mathcal{G}_t) \mapsto \mathcal{V}_t,
        \label{eq:panoptic_occupancy_prediction_mapping}
    \end{equation}
    where $f$ is a camera-centric panoptic occupancy prediction model.
    To reduce storage cost and loading latency, the canonical form of panoptic occupancy is defined as a sparse occupied-voxel set:
    \begin{equation}
        \mathcal{V}_t
        =
        \{(\boldsymbol{v}_{t,n},c_{t,n},i_{t,n})
        \mid n=1,\ldots,N_t\},
        \label{eq:sparse_panoptic_occupancy_set}
    \end{equation}
    where $\boldsymbol{v}_{t,n}\in\mathbb{Z}^{3}$ stores discrete voxel indices at timestamp $t$, $c_{t,n}\in\mathcal{C}$ denotes the semantic category, and $i_{t,n}\in\mathbb{N}_{0}$ denotes the instance identity.
    Let $\mathcal{C}_{\mathrm{th}}$ and $\mathcal{C}_{\mathrm{st}}$ be the disjoint thing and stuff subsets of $\mathcal{C}$.
    Stuff voxels have $i_{t,n}=0$, whereas voxels of the same thing object share a positive instance identity.
    Without loss of generality, the semantics and instance identity of each voxel are encoded by the following panoptic label:
    \begin{equation}
        p_{t,n} = \beta\,c_{t,n} + i_{t,n},
        \label{eq:panoptic_label_encoding}
    \end{equation}
    where $\beta$ is a predefined encoding base larger than the maximum instance identity, so that each $p_{t,n}$ uniquely recovers the semantic--instance pair.
    Beyond label encoding, each frame is stored as a sparse occupied-voxel list together with a packed dense valid mask and grid metadata.
    Voxel coordinates and panoptic labels are serialized in Fortran order and written as compressed NPZ archives, which substantially reduces storage footprint and loading latency.
    This canonical representation provides a unified interface for dataset construction, training, and evaluation across all PanoBench subsets and converted public occupancy annotations.

    \subsection{PanoBench Overview}
    \label{subsec:panobench_overview}

	\begin{table*}[t]
		\centering
		\settablefont
		\caption{Comparison of dataset statistics between PanoBench and representative occupancy prediction benchmarks.}
		\label{tab:dataset_comparison}
		\setlength{\tabcolsep}{3.5pt}
		\begin{tabular}{lccccccccc}
			\toprule
			Dataset
			& Data Source
			& View
			& Sequences
			& Frames
			& Classes
			& Voxel Size (m)
			& Volume Size (m)
			& Instance
			& Valid Masks \\
			\midrule
			SemanticKITTI~\cite{semantickitti2019behley} & Real            & Forward           & 22  & 43,000  & 28 & 0.2           & [51.2, 51.2, 6.4]   & $\times$     & $\times$ \\
			SSCBench-KITTI360~\cite{sscbench2024li}      & Real            & Forward           & 9   & 12,865  & 19 & 0.2           & [51.2, 51.2, 6.4]   & $\times$     & $\times$ \\
			Occ3D-nuScenes~\cite{occ3d2023tian}          & Real            & Surround          & 850 & 34,149  & 17 & 0.4           & [80.0, 80.0, 6.4]   & $\times$     & $\checkmark$ \\
			SurroundOcc~\cite{surroundocc2023wei}        & Real            & Surround          & 850 & 34,149  & 17 & 0.5           & [100.0, 100.0, 8.0] & $\times$     & $\times$ \\
			CarlaSC~\cite{motionsc2022wilson}            & Synthetic       & Surround          & 24  & 43,200  & 11 & 0.2, 0.4      & [51.2, 51.2, 3.0]   & $\times$     & $\times$ \\
			CoHFF~\cite{collaborative2024song}           & Synthetic       & Surround          & 73  & 11,464  & 12 & 0.4           & [40.0, 40.0, 3.2]   & $\times$     & $\times$ \\
			\midrule
			\textbf{PanoBench-CarlaOcc}                  & Synthetic       & Forward, Surround & 104 & 104,000 & 30 & 0.1, 0.2, 0.4 & [51.2, 51.2, 6.4]   & $\checkmark$ & $\checkmark$ \\
			\textbf{PanoBench-nuScenes}                  & Real            & Surround          & 850 & 34,149  & 17 & 0.2, 0.4      & [80.0, 80.0, 6.4]   & $\checkmark$ & $\checkmark$ \\
			\textbf{PanoBench-KITTI360}                  & Real            & Forward           & 9   & 76,473  & 19 & 0.2           & [51.2, 51.2, 6.4]   & $\checkmark$ & $\checkmark$ \\
			\textbf{PanoBench (total)}                   & Synthetic, Real & Forward, Surround & 963 & 214,622 & -- & 0.1, 0.2, 0.4 & --                  & $\checkmark$ & $\checkmark$ \\
			\bottomrule
		\end{tabular}
	\end{table*}
    PanoBench is a synthetic and real-world benchmark family for camera-centric panoptic occupancy prediction.
    It contains one synthetic subset, PanoBench-CarlaOcc, and two real-world subsets, PanoBench-nuScenes and PanoBench-KITTI360.
    According to the data domain, these subsets form a synthetic branch and a real-world branch, and they are further divided into diverse variants by camera view and voxel resolution.
    All variants share the canonical occupancy form, valid-mask representation, and model interface defined in Sects.~\ref{subsec:task_formulation} and~\ref{subsec:valid_mask}, enabling fair comparisons of semantic occupancy prediction, panoptic occupancy prediction, 4D occupancy forecasting, and sim-to-real transfer under a common protocol.

    PanoBench-CarlaOcc provides the synthetic branch.
    It contains 104 driving sequences and 104,000 frames collected from eight CARLA towns, covering 30 semantic classes and temporally consistent instance identities.
    To support controlled studies, it provides both forward-view and surround-view variants over a $51.2\times51.2\times6.4$\,m occupancy volume at voxel resolutions of 0.1/0.2/0.4\,m.
    The forward-view variant matches the limited-view setting of classical scene completion, whereas the surround-view variant supports full camera-centric occupancy prediction.
    Because its labels are derived from simulated meshes, PanoBench-CarlaOcc supplies complete geometry and serves as the controlled source domain for resolution analysis and sim-to-real experiments.

    The real-world branch contains PanoBench-nuScenes and PanoBench-KITTI360.
    PanoBench-nuScenes follows the surround-view camera setting of nuScenes~\cite{nuscenes2020caesar} and provides 34,149 frames from 850 scenes, 17 semantic classes, and 0.2/0.4\,m panoptic occupancy labels within an $80\times80\times6.4$\,m volume.
    PanoBench-KITTI360 follows the forward-view setting of KITTI-360~\cite{kitti3602023liao} and provides 76,473 frames from nine sequences, 19 semantic classes, and 0.2\,m labels within a $51.2\times51.2\times6.4$\,m volume.
    Unlike the synthetic subset, these real-world subsets are reconstructed from multi-view RGB images, semantic LiDAR observations, HD map evidence, and 3D object annotations, while further providing complete foreground assets and persistent instance labels.
    Table~\ref{tab:dataset_comparison} summarizes the three PanoBench subsets and compares them with other representative occupancy prediction benchmarks.

    \subsection{PanoBench-CarlaOcc: A Synthetic Panoptic Occupancy Dataset}
    \label{subsec:panobench_carlaocc}

    PanoBench-CarlaOcc is built upon our previously introduced dataset CarlaOcc~\cite{carlaocc2026feng}.
    Unlike public occupancy prediction benchmarks reconstructed from incomplete sensor observations, it obtains complete scene geometry and object identities directly from the simulation environment.
    The original simulation and annotation pipeline is detailed in our previous work~\cite{carlaocc2026feng}, while this study summarizes the dataset construction and its integration into PanoBench.

    \subsubsection{Data Collection}
    The synchronized multi-modal driving sequences are collected using CARLA~\cite{carla2017dosovitskiy} with Unreal Engine~5~\cite{unrealengine5}.
    The sensor rig follows the camera--LiDAR layout of the KITTI-360~\cite{kitti3602023liao} dataset and adds rear-facing cameras for complete surround coverage.
    Six cameras are configured to simultaneously render RGB, depth, and semantic images at a resolution of $1{,}408\times376$ pixels, together with a 32-channel LiDAR of 80\,m range.
    All sensors are calibrated on the ego vehicle and synchronized at 20\,Hz.
    To increase scene diversity, the sequences cover different road layouts and architectural styles at four traffic-density levels.
    The dataset further provides 3D bounding boxes, velocities of traffic participants, and more than 40K object mesh assets.

    \subsubsection{Mesh-Based Scene Reconstruction}
    Panoptic occupancy ground truth of CarlaOcc is reconstructed from simulated meshes rather than accumulated LiDAR points.
    For each frame, the meshes of static background that intersect the target occupancy volume are first retrieved and transformed into the ego-vehicle coordinate system.
    Subsequently, the dynamic foreground instances are incorporated using their simulated poses and instance identities.
    For non-rigid instances such as pedestrians,
    their best-fitting meshes are retrieved from the simulator's standard-mesh library according to skeleton similarities.
    The whole process produces a frame-level mesh-based scene representation containing the background layout, movable objects, and deformable participants in a geometrically consistent coordinate system.

    \subsubsection{Panoptic Occupancy Generation}
    We voxelize the reconstructed scene meshes and assign each occupied voxel its semantic and instance labels.
    Spatially overlapped components are merged in a topology-aware order that preserves thin ground structures and prevents stuff regions from overwriting foreground instances.
    Stuff meshes of the same category are consolidated before voxelization, whereas thing instances remain separated.
    The dense labels are then converted to the canonical sparse form in Sect.~\ref{subsec:task_formulation}.
    Ground categories are regularized as single-layer surfaces to avoid redundant vertical thickness and to match the real-world reconstruction protocol.

    \begin{figure*}[!t]
        \centering
        \includegraphics[width=0.99\textwidth]{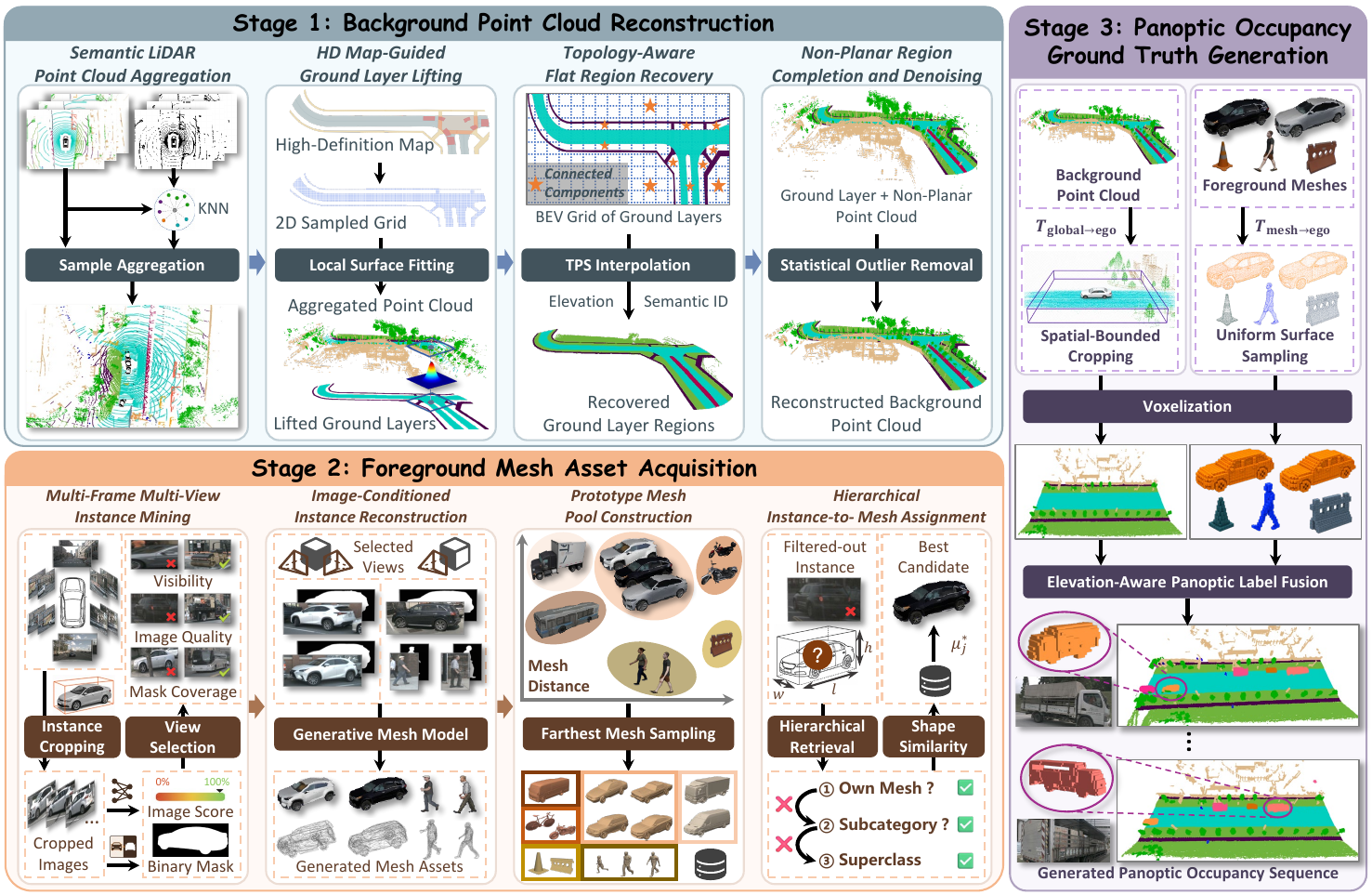}
        \caption{Real-world panoptic occupancy reconstruction pipeline. Static background geometry is recovered by semantic LiDAR aggregation, HD map-guided ground lifting, topology-aware flat region recovery, and geometry-preserving processing of non-planar structures. For foreground thing instances, PanoBench mines high-quality multi-frame multi-view image crops, reconstructs complete mesh assets, selects a diverse prototype pool, and assigns a compatible mesh to each annotated instance. The aligned foreground meshes and scene-level background are voxelized and fused into sparse panoptic occupancy labels, from which temporally aware camera-centric valid masks are generated.}
        \label{fig.real_world_pipeline}
    \end{figure*}

    \subsection{Real-World Panoptic Occupancy Reconstruction}
    \label{subsec:real_world_reconstruction}
    Although synthetic data provides high-fidelity and physically consistent panoptic annotations, its scene diversity remains limited, and an unavoidable simulation-to-real domain gap persists.
    Constructing panoptic occupancy annotations for real-world scenes nevertheless remains challenging, as sensor observations are sparse, noisy, and incomplete.
    A single LiDAR sweep captures only visible surfaces, whereas conventional multi-frame aggregation adopted in previous benchmarks~\cite{semantickitti2019behley,openoccupancy2023wang,occ3d2023tian,surroundocc2023wei,motionsc2022wilson,sscbench2024li} can introduce ghosting artifacts from dynamic objects and yield fragmented structures in sparsely observed distant regions.

    To address these challenges, we organize real-world panoptic occupancy annotation into three coordinated stages: background point cloud reconstruction, foreground mesh asset acquisition, and panoptic occupancy ground truth generation.
    As shown in Fig.~\ref{fig.real_world_pipeline}, the proposed pipeline enables static scene structures and dynamic instances to be reconstructed from complementary sources of evidence and subsequently integrated into a unified panoptic occupancy volume, providing geometrically complete, temporally consistent, and instance-aware panoptic occupancy annotations for real-world driving datasets~\cite{nuscenes2020caesar,kitti3602023liao}.

    \subsubsection{Background Point Cloud Reconstruction}
    \label{subsubsec:bg_recons}

    \noindent\textbf{Semantic LiDAR Point Cloud Aggregation.}
    Points of foreground instances are first removed from each LiDAR sweep to prevent motion-induced trailing artifacts, and the remaining static observations are aggregated across each driving sequence to form a scene-level representation.
    The raw point clouds are transformed into a common coordinate system using calibrated ego poses, and their semantic labels are retained during accumulation.
    To densify sparsely observed non-planar structures, unlabeled LiDAR sweeps inherit semantics from their nearest labeled observations subject to a distance constraint.
    This process increases spatial coverage beyond a single frame and provides the primary geometric evidence for static background structures.

    \noindent\textbf{HD Map-Guided Ground Layer Lifting.}
    Although multi-frame LiDAR aggregation expands the spatial coverage of the scene, ground-layer measurements remain discontinuous and fragmented because LiDAR point density decays rapidly with range.
    Moreover, regions missed by the LiDAR sensor due to occlusion are indistinguishable from truly empty space in the voxel space, causing physically continuous surfaces to be incorrectly truncated in the reconstructed background.
    Furthermore, ground-layer geometry can be sufficiently represented by a single lifted surface layer, since additional vertical thickness increases storage cost without providing meaningful supervision for camera-centric scene reconstruction.
    Based on these observations, we utilize HD map polygons as horizontal priors and aggregated LiDAR points as vertical evidence to lift 2D HD map regions into 3D ground layers.

    Let the aggregated LiDAR observations of the $\ell$-th ground layer be $\{(\boldsymbol{u}_i,z_i) \mid i=1,\ldots,N_{\ell}\}$, where $\boldsymbol{u}_i=(x_i,y_i)^{\top}$ is the horizontal position and $z_i$ is the measured height in the global coordinate system.
    For each ground layer $\ell$, its 2D polygon set $\Omega_{\ell}=\{\Omega_{\ell,q}\}_{q=1}^{R_\ell}$ is retrieved from the HD map annotations, and the enclosed valid region is uniformly sampled on a 2D grid, expressed as follows:
    \begin{equation}
        \mathcal{U}_{\ell}
        =
        \left\{\boldsymbol{u} \mid
            \boldsymbol{u}\in
            \cup_{q=1}^{R_\ell}\operatorname{Int}(\Omega_{\ell,q})
            \cap \rho\mathbb{Z}^{2}\right\},
        \label{eq:ground_sampling}
    \end{equation}
    where $\operatorname{Int}(\cdot)$ denotes the planar interior enclosed by a map polygon, and $\rho$ is the sampling interval, selected as half of the voxel size to avoid voxelization aliasing.
    For each sampled ground-layer point $\boldsymbol{u}\in\mathcal{U}_{\ell}$, its elevation is estimated from neighboring LiDAR ground observations $\mathcal{N}(\boldsymbol{u})$ via distance-weighted local surface fitting:
    \begin{equation}
        \boldsymbol{\theta}_{\boldsymbol{u}}^{*}
        =
        \arg\min_{\boldsymbol{\theta}}
        \sum_{\boldsymbol{u}_i\in\mathcal{N}(\boldsymbol{u})}
        \omega_i
        \left(
            [\boldsymbol{u}_i^{\top},1]\boldsymbol{\theta}
            -
            z_i
        \right)^2,
        \label{eq:ground_local_surface_fitting}
    \end{equation}
    where $\omega_i=\exp\!\left(-\|\boldsymbol{u}-\boldsymbol{u}_i\|_2^2/(2\sigma^2)\right)$ is a Gaussian-decayed distance weight with bandwidth $\sigma$, which yields a smooth and accurate estimate of the grid-point elevation.
    The lifted height is then given by:
    \begin{equation}
        H_{\boldsymbol{u}}
        =
        [\boldsymbol{u}^{\top},1]\boldsymbol{\theta}_{\boldsymbol{u}}^{*}.
        \label{eq:ground_lifted_height}
    \end{equation}
    The resulting lifted samples are assigned semantic labels according to their corresponding map layers and aggregated into a scene-level ground-surface point cloud.

    \noindent\textbf{Topology-Aware Flat Region Recovery.}
    Despite providing strong horizontal priors, HD map polygons may contain unlabeled holes or unobserved flat regions, even when the underlying ground surface is physically continuous.
    Directly voxelizing such polygons would imprint these annotation defects onto the occupancy volume as artificial discontinuities.
    We therefore introduce a topology-aware recovery strategy that reasons jointly about spatial continuity and the semantic adjacency of neighboring map layers.
    A connected-component analysis is first performed in the BEV grid to identify candidate flat regions that are spatially adjacent to the reconstructed ground layers but lack semantic map priors.
    To verify that these candidates belong to ground surfaces, the consistency between their local normals and the vertical direction is validated.
    Subsequently, the boundary point sets shared with the known ground regions are extracted and employed with a Thin-Plate Spline (TPS) interpolation~\cite{bookstein1989tps} to construct a continuous 2D elevation field over the candidate domain.
    This formulation naturally preserves gentle slopes and curvature without introducing artificial steps.
    Dense query points are then sampled within the BEV-convex silhouette of the candidate region, their fitted elevations are computed via the interpolated field, and semantic labels are assigned via nearest-neighbor voting~\cite{cover1967nearest} from the aggregated point cloud.
    To prevent unbounded extrapolation, we constrain the propagation range within a limited buffer distance from the known seeds, leaving distant unobserved terrains unfilled to avoid hallucination.
    This procedure repairs map-induced artifacts while preserving mutually exclusive semantic boundaries and avoiding unnecessary modifications to reliably measured geometry.

    \noindent\textbf{Non-Planar Region Completion and Denoising.}
    For non-planar static structures such as vegetation clusters and man-made objects, aggregated point clouds suffer from severe sparsity and noise due to LiDAR multipath reflections, occlusions, and ranging errors.
    Statistical outlier removal is therefore applied: a point is discarded if its average distance to neighboring points is significantly larger than the scene-level distribution of this distance.
    This step suppresses isolated measurement noise and floating artifacts while preserving locally dense structures.
    A semi-automated interface then supports visual inspection and local correction of the background point cloud, using synchronized camera images and projected LiDAR points as reference.
    The combination of statistical filtering and targeted refinement yields clean, high-quality background point clouds that serve as reliable geometric priors for the final occupancy ground truth.

    \subsubsection{Foreground Mesh Asset Acquisition}
    \label{subsubsec:fg_recons}
    Foreground objects require instance-level, complete geometry rather than class-level surface reconstruction.
    Existing real-world occupancy prediction benchmarks typically reconstruct foreground objects by directly aggregating semantic LiDAR observations over time~\cite{semantickitti2019behley,surroundocc2023wei,occ3d2023tian,openoccupancy2023wang,sscbench2024li}.
    This observation-driven strategy is inherently limited for dynamic instances, since temporal aggregation can introduce motion-induced ghosting and significant geometric distortion.
    Moreover, the aggregated LiDAR sweeps capture only visible surfaces, which cannot recover the occluded portions of an object.
    In contrast, we decouple foreground reconstruction from sparse geometric observations and formulate it as image-conditioned mesh asset generation, in which off-the-shelf generative 3D foundation models~\cite{trellis2024xiang,trellis22025xiang,sam3d2026chen} infer complete instance-level geometry from sparse visual observations.

    \noindent\textbf{Multi-Frame Multi-View Instance Mining.}
    We first retrieve all target samples and collect annotated instances whose 3D boxes intersect the occupancy volume.
    For each candidate instance, its full temporal trajectory is traversed, its 3D bounding box is projected into the synchronized multi-view image sequences, and a tightly padded patch is cropped from each view in which the instance is sufficiently visible and not severely truncated.
    Each crop is then assessed from two complementary aspects: an image-quality evaluator (\eg, MANIQA~\cite{maniqa2022yang}) measures perceptual sharpness and exposure fidelity, while a text-prompted segmenter (\eg, SAM~3~\cite{sam32025carion}) produces an instance mask that provides foreground-coverage and occlusion cues.
    Views that jointly satisfy image-quality, mask-coverage, and multi-camera constraints are selected as candidate views for mesh reconstruction.
    When multiple observations of the same instance are available, the views with stronger instance support are retained to improve mesh reconstruction quality.
    This process filters out heavily occluded, motion-blurred, and poorly segmented observations, yielding a compact set of clean, viewpoint-diverse inputs per instance.

    \noindent\textbf{Image-Conditioned Instance Reconstruction.}
    After view selection, each retained instance is represented by a compact set of crop-mask pairs.
    These RGB crops and their binary masks are fed into a generative mesh model (\eg, MV-SAM3D~\cite{li2026mv}), where the masks provide explicit foreground support and exclude background content.
    Instances with multiple valid observations are reconstructed through the multi-view branch, while rare single-view cases fall back to the single-view branch.
    The decoded meshes and Gaussians are jointly baked into a textured GLB model via differentiable rendering, after which an adaptive quadric simplification is applied to reduce polygon count while preserving silhouette fidelity.
    The resulting meshes serve as reusable, scene-agnostic geometric prototypes for other instances.

    \noindent\textbf{Prototype Mesh Pool Construction.}
    The set of reconstructed assets is both redundant and incomplete: most reconstructed meshes are near-duplicates, whereas the view-selection criteria may leave many in-range objects without a mesh of their own.
    We therefore curate a compact prototype mesh pool that preserves both semantic coverage and intra-class shape diversity, so that later assignment can supply a valid mesh to every in-range instance.

    The reconstructed meshes are first grouped by the dataset taxonomy, and then centered and normalized by the bounding-box diagonal to remove the absolute scale.
    Within each category, surface points are uniformly sampled from each mesh, and the pairwise discrepancy is measured by the following symmetric Chamfer distance~\cite{fan2017pointset}:
    \begin{equation}
        \begin{split}
            s(\mu,\nu)
            ={}&
            \frac{1}{|\mathcal{S}_{\mu}|}
            \sum_{\boldsymbol{x}\in\mathcal{S}_{\mu}}
            \min_{\boldsymbol{y}\in\mathcal{S}_{\nu}}
            \|\boldsymbol{x}-\boldsymbol{y}\|_2
            \\
            &+
            \frac{1}{|\mathcal{S}_{\nu}|}
            \sum_{\boldsymbol{y}\in\mathcal{S}_{\nu}}
            \min_{\boldsymbol{x}\in\mathcal{S}_{\mu}}
            \|\boldsymbol{y}-\boldsymbol{x}\|_2,
        \end{split}
        \label{eq:prototype_mesh_distance}
    \end{equation}
    where $\mathcal{S}_{\mu}$ and $\mathcal{S}_{\nu}$ are the sampled point sets of meshes $\mu$ and $\nu$.
    Inspired by farthest point sampling~\cite{qi2017pointnetplusplus}, we introduce \textit{Farthest Mesh Sampling} to select representative yet complementary prototypes.
    For category $c$, with candidate mesh set $\mathcal{R}_{c}$ and selected mesh set $\mathcal{B}_{c}$, the following mesh asset is iteratively added to the prototype pool:
    \begin{equation}
        \mu^{*}
        =
        \arg\max_{\mu\in\mathcal{R}_{c}\setminus\mathcal{B}_{c}}
        \min_{\nu\in\mathcal{B}_{c}}
        s(\mu,\nu).
        \label{eq:farthest_mesh_sampling}
    \end{equation}
    The resulting prototype mesh pool removes near-duplicates while retaining the intra-class variations required by hierarchical assignment.

    \noindent\textbf{Hierarchical Instance-to-Mesh Assignment.}
    While strict view selection enhances mesh reconstruction quality, it also prevents a considerable number of foreground instances from being successfully reconstructed.
    We therefore assign a mesh asset from the prototype mesh pool to every required foreground instance using deterministic hierarchical retrieval.
    The assignment first uses the instance's own reconstructed mesh if it exists.
    Otherwise, it progressively relaxes the scope from the same subcategory to the semantic superclass when no finer-grained asset is available.

    Let $\boldsymbol{b}_j=(w_j,l_j,h_j)^\top$ denote the annotated dimensions of target instance $j$, and let $\boldsymbol{b}_{\mu}$ denote the dimensions of the source instance associated with prototype mesh $\mu$.
    Within the selected candidate-index set $\mathcal{A}(j)$, the most compatible prototype for target instance $j$ is chosen as follows:
    \begin{equation}
        \mu_j^{*}
        =
        \mathop{\operatorname{arg\,lexmin}}_{\mu\in\mathcal{A}(j)}
        \Bigl(
        \Bigl\|
        \frac{\boldsymbol{b}_j}{\|\boldsymbol{b}_j\|_2}
        -
        \frac{\boldsymbol{b}_{\mu}}{\|\boldsymbol{b}_{\mu}\|_2}
        \Bigr\|_2,\,
        \|\boldsymbol{b}_j-\boldsymbol{b}_{\mu}\|_2
        \Bigr),
        \label{eq:instance_mesh_assignment}
    \end{equation}
    where $\operatorname{arg\,lexmin}$ minimizes the ordered pair lexicographically.
    Proportional shape compatibility is therefore the primary criterion, while absolute dimensional similarity serves as a deterministic tie-breaker among proportionally equivalent candidates.
    The resulting instance-to-mesh assignment produces a semantically valid mesh for every in-range foreground instance and records the selected prototype for subsequent panoptic occupancy generation.

    \subsubsection{Panoptic Occupancy Ground Truth Generation}
    \label{subsubsec:pano_occ_gt_gen}
    The preceding stages yield a globally consistent semantic point cloud for background stuff classes and an instance-specific mesh assignment for each annotated thing object.
    At the final stage, both representations are instantiated in a common frame-aligned voxel grid and assembled into per-frame sparse panoptic occupancy annotations.

    The scene-level background point cloud is first transformed into the current ego frame.
    Spatial-bounded cropping is adopted to retain only points inside the target occupancy range, and the remaining semantic points are quantized on the predefined voxel lattice.
    The resulting background candidates keep their semantic labels with an instance identity of zero.
    For each foreground thing instance, the assigned prototype mesh is scaled and posed according to the annotated 3D bounding box and then transformed into the current ego frame.
    Surface points are uniformly sampled from the aligned mesh and voxelized on the same lattice as the background, preserving the semantic category and persistent instance identity of the object.

    A direct merge of the voxelized background and foreground may assign multiple labels to the same voxel, since different stuff classes can overlap after point-cloud aggregation and foreground instances can intersect with the reconstructed background.
    We therefore introduce Elevation-Aware Panoptic Label Fusion to resolve these conflicts in a unified manner.
    A category-level elevation-priority lookup table is predefined according to the real-world topological relationships among semantic classes.
    For each frame, all labeled voxel candidates are assigned priorities from this table and then sorted together with their world-space elevations.
    Candidates are written into the occupancy grid in increasing elevation-priority order: when several labels claim the same voxel, the higher-priority candidate is kept.
    This prevents ground and other low structures from overwriting higher surfaces or foreground objects.
    This process produces a per-frame panoptic occupancy ground truth containing occupied voxel coordinates, semantic categories, and persistent instance identities.

    \begin{figure}[!t]
        \centering
        \includegraphics[width=0.99\columnwidth]{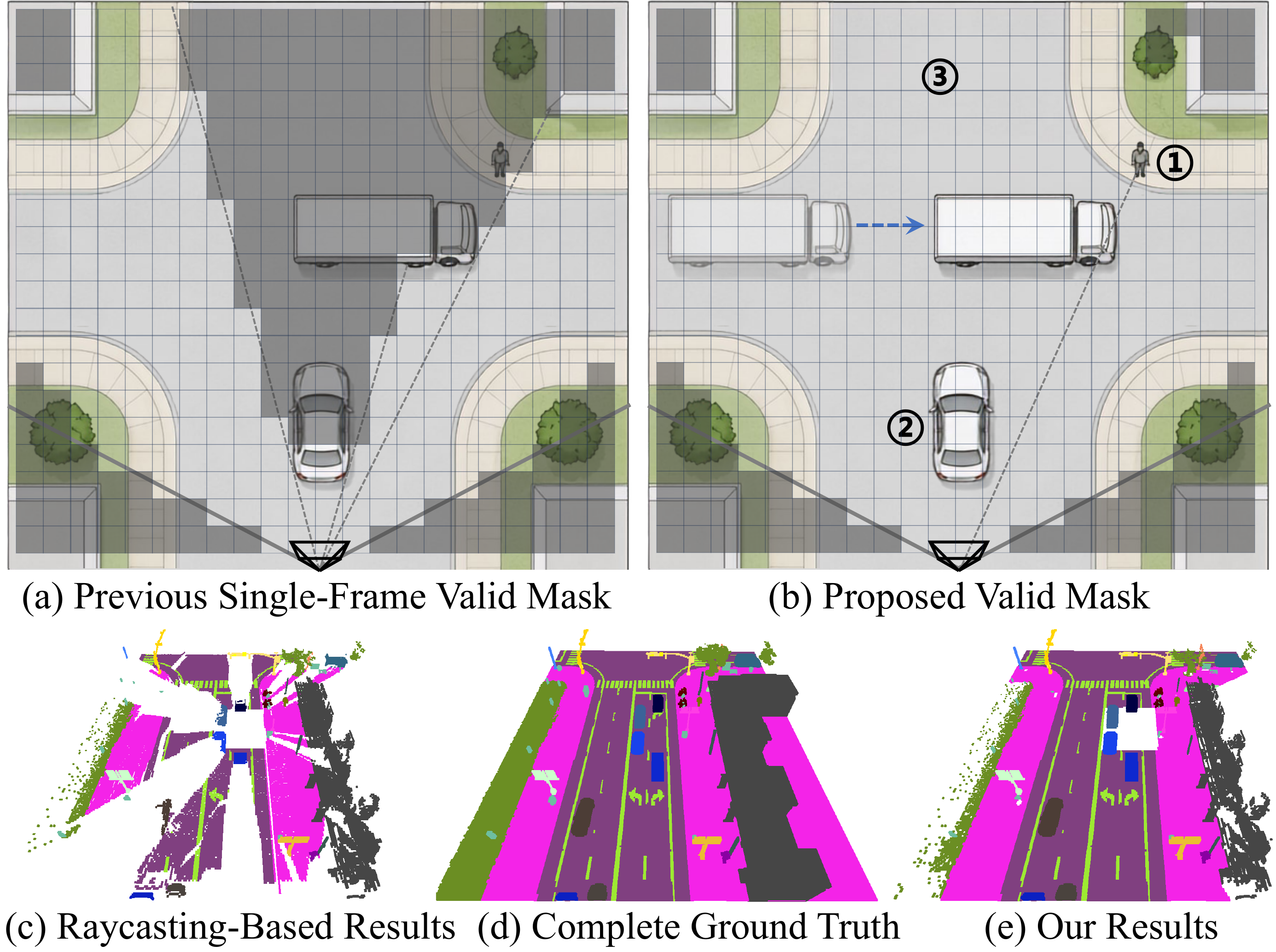}
        \caption{Comparison of prior valid masks and the proposed valid masks.
            (a)~Previous methods generate a valid mask for each single frame, in which camera rays leave large occluded regions invalid (gray).
            (b)~The proposed temporal valid mask keeps three classes of regions under evaluation:
                {\textcircled{\tiny{$\mathbf{1}$}}}~a currently occluded but previously observed instance, recovered by aggregating visibility over a window of~$T$ frames;
                {\textcircled{\tiny{$\mathbf{2}$}}}~a partially observed instance, completed as a whole once any of its voxels is visible; and
                {\textcircled{\tiny{$\mathbf{3}$}}}~stuff behind foreground instances, which remains valid to avoid occlusion-induced fragments.
            (c)--(e)~Qualitative comparison of the raycasting-based mask, the complete ground truth, and the proposed valid mask on PanoBench-CarlaOcc.}
        \label{fig.valid_mask}
    \end{figure}

    \subsection{Valid Mask Generation}
    \label{subsec:valid_mask}
    To avoid scoring regions that the cameras cannot observe, existing occupancy prediction benchmarks commonly define valid masks from current-frame sensor evidence.
    Occ3D~\cite{occ3d2023tian} provides camera visibility masks, while SSCBench~\cite{sscbench2024li} inherits voxelized invalid masks from scene completion benchmarks~\cite{semantickitti2019behley,kitti3602023liao}.
    These protocols are effective for evaluating visible surfaces, but they tend to encourage occupancy prediction to degenerate into depth estimation with semantic segmentation, where models are rewarded for the currently visible surfaces rather than for complete 3D structures.
    The problem is more severe for panoptic occupancy, because a thing object must be reconstructed as a whole instance, not as a visible fragment.
    Moreover, most existing valid masks are generated independently for each target frame, which ignores objects that were observed in the recent past but occluded at present.
    Consequently, this design understates the role of temporal observations and cannot evaluate whether a model keeps instance identity and completes an object once it is occluded.

    Motivated by these observations, we construct a temporally aware valid mask $\boldsymbol{M}_t\in\{0,1\}^{X\times Y\times Z}$ from the target occupancy $\mathcal{V}_t$ in a window of $T$ frames ending at $t$.
    The valid mask comprises an instance-level thing mask $\boldsymbol{M}_t^{\mathrm{th}}$, a voxel-level stuff mask $\boldsymbol{M}_t^{\mathrm{st}}$, and a visible free-space mask $\boldsymbol{M}_t^{\mathrm{free}}$, all of the same shape. As shown in Fig.~\ref{fig.valid_mask}, the proposed valid mask produces more reasonable and comprehensive evaluation regions than the previous methods.

    \noindent\textbf{Voxel Visibility.}
    For each camera $k$ at timestamp $\tau$, voxels outside the camera view are first filtered out.
    Each retained voxel is then tested by ray marching through the occupancy grid from the camera center.
    The voxel is visible only if the ray reaches it before a blocking occupied cell.
    Pixel-driven alternatives such as image-space z-buffering or per-pixel raycasting are avoided, because the mismatch between pixel and voxel sampling rates may produce stripe artifacts at long range.
    Voxel visibility is aggregated by a union over the temporal window, expressed as follows:
    \begin{equation}
        \chi_t(\boldsymbol{v})
        =
        1 - \prod_{\tau=t-T+1}^{t}\prod_{k=1}^{K}
        \bigl(1-\chi_{\tau}^{k}(\boldsymbol{v})\bigr),
        \label{eq:historical_voxel_visibility}
    \end{equation}
    where $\chi_{\tau}^{k}(\boldsymbol{v})=1$ when the ray from camera~$k$ at time~$\tau$ reaches $\boldsymbol{v}$ before any blocking cell,
    and $\chi_t(\boldsymbol{v})=1$ means that at least one camera in the window observes $\boldsymbol{v}$.

    \noindent\textbf{Thing, Stuff, and Free Space.}
    Thing classes require complete instance geometry, so visibility is lifted from voxels to instances.
    Let $\mathcal{Q}_{t,j}^{\mathrm{th}}$ be the voxel set of thing instance $j$ at timestamp $t$, and $\mathcal{Q}_t^{\mathrm{st}}$ be the voxel set of all stuff classes.
    The instance is valid once any of its voxels is temporally visible:
    \begin{equation}
        \boldsymbol{M}_{t}^{\mathrm{th}}(\boldsymbol{v})
        =
        \begin{cases}
        \mathbb{I}\!\left[
                        \exists\,\boldsymbol{u}\in\mathcal{Q}_{t,j}^{\mathrm{th}}:
                        \chi_t(\boldsymbol{u})=1
        \right],
        &
        \boldsymbol{v}\in\mathcal{Q}_{t,j}^{\mathrm{th}},
        \\
        0,
        &
        \text{otherwise,}
        \end{cases}
        \label{eq:thing_instance_mask}
    \end{equation}
    where $\mathbb{I}\left[\cdot\right]$ denotes the indicator function.
    Once an object has been partially observed in the window, its full occupancy enters the evaluation region.
    This asks a temporal model to keep the instance and complete its shape, rather than to predict only the currently visible fragment.

    Stuff classes such as road and sidewalk lack well-defined instance boundaries, so instance-level expansion would flood large amorphous regions.
    A stuff voxel is therefore valid only if it itself is visible:
    \begin{equation}
        \boldsymbol{M}_{t}^{\mathrm{st}}(\boldsymbol{v})
        =
        \begin{cases}
        \chi_t(\boldsymbol{v}),
        &
        \boldsymbol{v}\in\mathcal{Q}_t^{\mathrm{st}},
        \\
        0,
        &
        \text{otherwise.}
        \end{cases}
        \label{eq:stuff_voxel_mask}
    \end{equation}
    Meanwhile, foreground things are not treated as occluders of background stuff, so that a vehicle does not invalidate the road surface behind or beneath it.

    Visible free space is also marked as valid during evaluation.
    For each in-frustum camera ray, empty voxels are traversed until the first occupied hit, and all traversed empty voxels are marked valid.
    This yields $\boldsymbol{M}_t^{\mathrm{free}}$, which confines free-space evaluation to the camera-observable volume instead of restricting evaluation to occupied voxels.

    The final mask combines the three tensors as follows:
    \begin{equation}
        \boldsymbol{M}_t
        =
        \boldsymbol{M}_t^{\mathrm{th}}
        \lor
        \boldsymbol{M}_t^{\mathrm{st}}
        \lor
        \boldsymbol{M}_t^{\mathrm{free}}.
        \label{eq:final_valid_mask}
    \end{equation}
    Compared with the previous frame-wise visibility mask, $\boldsymbol{M}_t$ keeps historically supported thing instances, scores stuff conservatively at the voxel level, and still restricts free space to rays that the cameras can actually see.

    \section{Experiments}
    \label{sec:experiments}

    \subsection{Experimental Setup}
    \label{subsec:experimental_setup}

    \subsubsection{Datasets and Benchmarks}
    \label{subsubsec:datasets_occupancy_benchmarks}
    The annotation quality of PanoBench is compared with that of public occupancy prediction benchmarks in three data domains, namely nuScenes~\cite{nuscenes2020caesar}, KITTI-360~\cite{kitti3602023liao}, and CARLA~\cite{carla2017dosovitskiy}, and existing camera-centric occupancy prediction methods are evaluated on the corresponding subsets PanoBench-nuScenes, PanoBench-KITTI360, and PanoBench-CarlaOcc.
    Occupancy prediction methods receive only RGB images together with camera intrinsics, extrinsics, and ego poses during training and inference.

    On nuScenes, PanoBench-nuScenes is compared with SurroundOcc~\cite{surroundocc2023wei}, Occ3D~\cite{occ3d2023tian}, and OpenOccupancy~\cite{openoccupancy2023wang}.
    Occupancy prediction methods use the six-camera surround-view setting.
    The default protocol uses a $0.4$\,m voxel size over an $80\times80\times6.4$\,m volume, with images resized to $256\times704$ pixels.
    The $0.2$\,m labels are used only in the annotation quality analyses in Sect.~\ref{subsec:dataset_annotation_quality}.
    Official scene-level training and validation splits are retained.

    On KITTI-360, PanoBench-KITTI360 is compared with SSCBench-KITTI360~\cite{sscbench2024li}.
    Occupancy prediction methods use the forward-view setting and the front camera only.
    The default protocol uses a $0.2$\,m voxel size over a $51.2\times51.2\times6.4$\,m volume, with images resized to $192\times640$ pixels.
    Sequence-level official partitions are used so that temporally adjacent frames do not cross splits.

    On CARLA, PanoBench-CarlaOcc is compared with CarlaSC~\cite{motionsc2022wilson}, and Occ3D-style and SurroundOcc-style labels are further reconstructed for the mesh-oracle study in Sect.~\ref{subsubsec:geometric_completeness}.
    Occupancy prediction methods use the six-camera surround-view setting.
    The default protocol uses a $0.4$\,m voxel size over a $51.2\times51.2\times6.4$\,m volume, with images resized to $192\times640$ pixels.
    The official split provided in CarlaOcc~\cite{carlaocc2026feng} is adopted for a fair comparison.
    For occupancy prediction, the $0.1$\,m and $0.2$\,m variants are used only in the multi-resolution analysis in Sect.~\ref{subsec:multi_resolution}.

    On each PanoBench subset, we evaluate semantic occupancy prediction, panoptic occupancy prediction, and 4D occupancy forecasting, where forecasting is evaluated at horizons of 1, 2, and 3\,s.

    \subsubsection{Benchmark Details and Metrics}
    \label{subsubsec:benchmark_details_metrics}
    The semantic benchmark includes GaussianFormer~\cite{gaussianformer2024huang}, OPUS~\cite{opus2024wang}, SparseOcc~\cite{sparseocc2024liu}, and Panoptic-FlashOcc~\cite{flashocc2024yu}.
    SparseOcc and Panoptic-FlashOcc additionally provide instance predictions for the panoptic benchmark.
    For 4D occupancy forecasting, SparseWorld~\cite{sparseworld2026dang} and SparseWorld-TC~\cite{sparseworldtc2025du} are benchmarked with the same dataflow and evaluation protocol.
    Unless otherwise stated, all methods on each subset share the same voxel size, spatial range, camera setting, input resolution, dataset split, and optimization budget.
    Eight NVIDIA RTX~4090 GPUs are used for training, and AdamW with an initial learning rate of $2\times10^{-4}$ is adopted.
    The learning rate follows a linear warm-up and a cosine annealing schedule over 20 epochs.
    The number of input frames is the only method-specific difference: multi-frame methods use eight frames including the target frame, whereas single-frame architectures use only the target frame.

    All metrics are computed under the proposed valid-mask protocol in Sect.~\ref{subsec:valid_mask}.
    Geometric IoU and semantic mIoU are reported for compatibility with prior work~\cite{tpvformer2023huang,opus2024wang}.
    RayIoU~\cite{sparseocc2024liu} and, for instance-aware methods, RayPQ~\cite{sparseocc2024liu} are evaluated at depth tolerances of 1, 2, and 4\,m, denoted by @1, @2, and @4, respectively.
    Forecasting methods are scored by geometric IoU and semantic mIoU at each horizon and by their averages over all horizons.
    Dataset-level confusion matrices and panoptic counts are accumulated before class averaging, and classes absent from an evaluation split are excluded.

    \begin{table}[t]
        \centering
        \settablefont
        \setlength{\tabcolsep}{4pt}
        \caption{Comparison of observation fidelity across occupancy benchmarks. }
        \label{tab:observation_fidelity_pixacc}
        \begin{tabular}{llcc}
            \toprule
            Domain & Dataset                                    & Voxel Size (m) & $\mathrm{PixAcc}$ (\%) \\
            \midrule
            \multirow{3}{*}{CARLA}
                   & \textbf{PanoBench-CarlaOcc}                & 0.1            & 91.6 \\
                   & \textbf{PanoBench-CarlaOcc}                & 0.2            & 88.8 \\
                   & \textbf{PanoBench-CarlaOcc}                & 0.4            & 86.4 \\
            \midrule
            \multirow{5}{*}{nuScenes}
                   & OpenOccupancy~\cite{openoccupancy2023wang} & 0.2            & 70.0 \\
                   & \textbf{PanoBench-nuScenes}                & 0.2            & \textbf{77.2} \\
                   & Occ3D~\cite{occ3d2023tian}                 & 0.4            & 69.2 \\
                   & \textbf{PanoBench-nuScenes}                & 0.4            & \textbf{73.0} \\
                   & SurroundOcc~\cite{surroundocc2023wei}      & 0.5            & 63.7 \\
            \midrule
            \multirow{2}{*}{KITTI-360}
                   & SSCBench-KITTI360~\cite{sscbench2024li}    & 0.2            & 69.9 \\
                   & \textbf{PanoBench-KITTI360}                & 0.2            & \textbf{81.5} \\
            \bottomrule
        \end{tabular}
    \end{table}

    \subsection{Occupancy Annotation Quality}
    \label{subsec:dataset_annotation_quality}

    Occupancy annotation quality is characterized by four distinct dimensions: \textbf{observation fidelity}, \textbf{geometric completeness}, \textbf{spatial continuity}, and \textbf{temporal consistency}.
    All cross-dataset comparisons align the region of interest, semantic taxonomy, and voxel resolution before evaluation.

    \subsubsection{Occupancy Observation Fidelity}
    \label{subsubsec:observation_fidelity}

\begin{figure*}[t]
        \centering
        \includegraphics[width=0.99\textwidth]{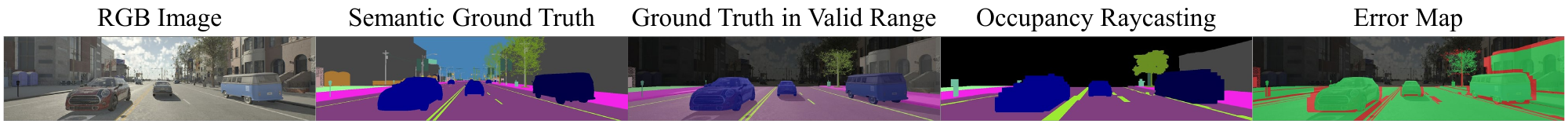}
        \caption{Illustration of observation fidelity. Occupancy is ray-cast into the camera view and compared with 2D semantic ground truth.}
        \label{fig.compare_observation_fidelity}
    \end{figure*}

    Observation fidelity measures whether an occupancy volume agrees with the corresponding 2D semantic evidence when rendered from the same camera view.
    As shown in Fig.~\ref{fig.compare_observation_fidelity}, each occupancy grid is ray-cast into the provided camera views to obtain rendered semantic maps, which are then compared with the 2D semantic ground truth.
    Pixel accuracy (PixAcc) is adopted as the observation fidelity score.
    Evaluation is restricted to pixels whose corresponding 3D points lie inside the occupancy ground-truth range.
    As shown in Table~\ref{tab:observation_fidelity_pixacc}, PanoBench annotations are consistently more observation-faithful than prior occupancy labels under matched voxel sizes.
    On nuScenes, PanoBench-nuScenes outperforms OpenOccupancy by $7.2$ points at $0.2$\,m and Occ3D by $3.8$ points at $0.4$\,m.
    On KITTI-360, PanoBench-KITTI360 reaches $81.5\%$ PixAcc at $0.2$\,m, outperforming SSCBench-KITTI360 by $11.6$ points.
    Smaller voxels also yield higher PixAcc, because finer grids better preserve object boundaries and thin structures under camera projection.
    As expected, synthetic annotations exhibit higher fidelity than real-world ones, benefiting from noise-free sensors and exact geometric ground truth in simulation.

    \begin{figure*}[!t]
        \centering
        \includegraphics[width=0.99\textwidth]{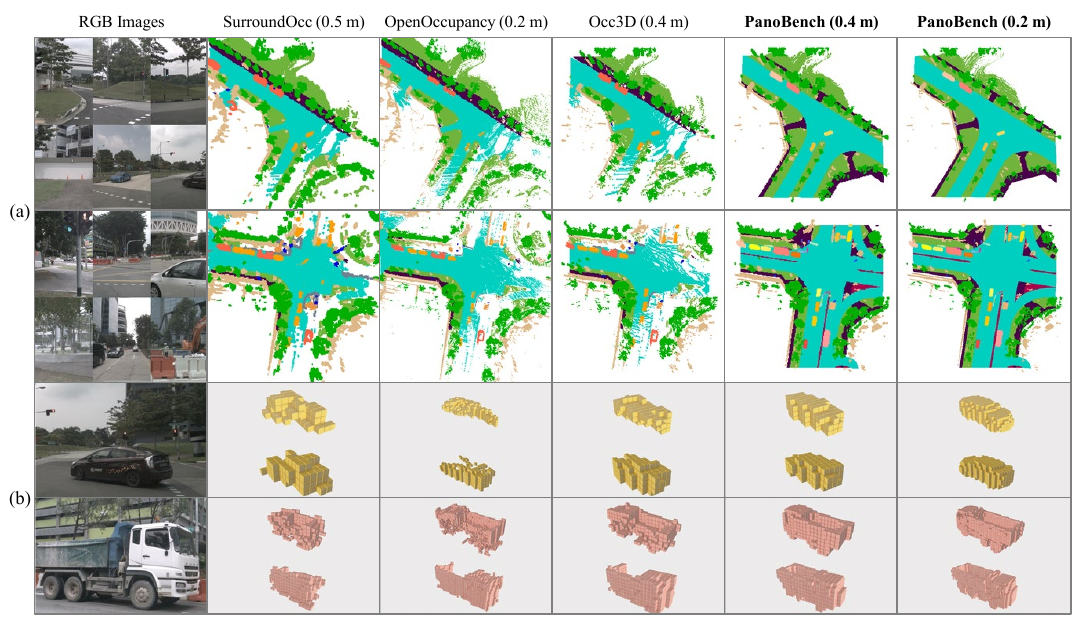}
        \caption{Comparison of geometric completeness on nuScenes. (a)~BEV views of two scenes. (b)~Two foreground instances, each rendered from two opposite viewpoints.}
        \label{fig.compare_dataset_quality}
    \end{figure*}

    \subsubsection{Geometric Completeness}
    \label{subsubsec:geometric_completeness}

    Geometric completeness measures whether occupancy annotations recover the full spatial extent of scene structures and object silhouettes, including occluded areas and regions with sparse sensor evidence.
    Because existing benchmarks reconstruct occupancy with different pipelines, their reconstruction results are compared both qualitatively on real data and quantitatively against a mesh-derived oracle on PanoBench-CarlaOcc.

    Fig.~\ref{fig.compare_dataset_quality} compares real-world occupancy annotations at both the scene level and the instance level under a shared semantic palette.
    SurroundOcc~\cite{surroundocc2023wei}, OpenOccupancy~\cite{openoccupancy2023wang}, and Occ3D~\cite{occ3d2023tian} exhibit broken ground layers and incomplete foreground objects.
    Their annotations are constructed mainly by aggregating and voxelizing LiDAR points, and therefore cannot recover geometry under occlusion or sparse observation.
    In contrast, PanoBench produces substantially more continuous backgrounds and more complete foreground shapes.

    A controlled oracle evaluation is further conducted on PanoBench-CarlaOcc.
    Real sensor observations cannot validate unobserved surfaces, so the Occ3D~\cite{occ3d2023tian} and SurroundOcc~\cite{surroundocc2023wei} annotation pipelines are compared with PanoBench under a complete mesh oracle.
    \texttt{Town10HD} is selected as the reconstructed scene, where the sensor input is frozen and
    occupancy is rebuilt with three pipelines that share the same poses, LiDAR sweeps, and evaluation grid.
    Occ3D aggregates multi-frame static semantic LiDAR and retains dynamic classes from the current frame only before voxelization.
    SurroundOcc uses the same aggregation and densifies the static cloud with Poisson reconstruction before voxelization.
    These are style-faithful reconstructions of the corresponding annotation protocols rather than exact replicas of the original toolchains.
    CarlaOcc labels are remapped into coarse classes to reduce taxonomy discrepancies across pipelines.
    We compare the three reconstructed annotations against the mesh oracle and report geometric IoU, semantic mIoU, and per-class IoU.
    As shown in Table~\ref{tab:oracle_pipeline_carla}, PanoBench achieves 58.9 IoU and 52.0 mIoU, outperforming the stronger Occ3D-style pipeline by 20.6 and 19.4 points, respectively.
    The largest gains occur on Vehicle, Pedestrian, Terrain, and Sidewalk.
    For example, Vehicle IoU increases from 5.9 to 59.1, since map-guided ground lifting and complete instance meshes recover volumes that LiDAR point clouds miss.
    Although the SurroundOcc-style pipeline densifies the static cloud through Poisson reconstruction, it improves neither IoU nor mIoU over the Occ3D-style pipeline, showing that denser surfaces alone do not imply more complete geometry and semantics.
    Building and Vegetation remain difficult for all three pipelines because distant facades and foliage are only partially observed.

\begin{figure*}[!t]
        \centering
        \includegraphics[width=0.99\textwidth]{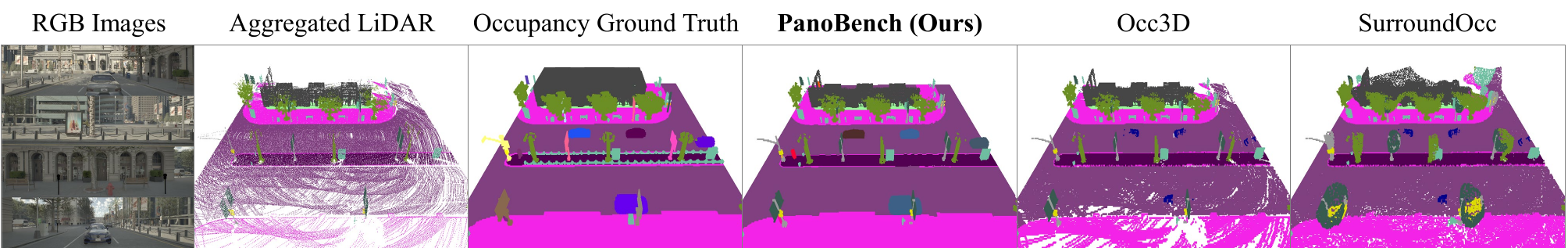}
        \caption{Comparison of occupancy annotation pipelines on PanoBench-CarlaOcc (\texttt{Town10HD}). The ground truth is voxelized from complete scene meshes, and Occ3D and SurroundOcc denote their style-faithful reimplementations.}
        \label{fig.oracle_pipeline}
    \end{figure*}

    Fig.~\ref{fig.oracle_pipeline} visualizes a representative frame.
    The Occ3D-style pipeline yields sparse, fragmented geometry with incomplete cars.
    The SurroundOcc-style pipeline recovers the ground layer better owing to Poisson reconstruction, but still produces deformed and erroneous geometry for complex structures such as buildings and traffic lights.
    In comparison, PanoBench recovers continuous ground and complete vehicles that closely match the mesh oracle.

    These results indicate that the incompleteness of previous LiDAR aggregation-based benchmarks is structural rather than incidental.
    PanoBench addresses this issue by exploiting geometric priors and reconstructing complete foreground assets, which substantially reduces missing geometry in occluded and sparsely observed regions.

    \begin{table*}[!t]
        \centering
        \settablefont
        \caption{Controlled oracle evaluation of occupancy annotation pipelines on PanoBench-CarlaOcc.}
        \label{tab:oracle_pipeline_carla}
        \setlength{\tabcolsep}{3.5pt}
        \begin{tabular}{l|cc|cccccccc}
            \toprule
            Method                                      & Geometric IoU & Semantic mIoU & Road          & Sidewalk      & Terrain       & Building      & Vegetation    & Object        & Vehicle       & Pedestrian \\
            \midrule
            Occ3D-style~\cite{occ3d2023tian}            & 38.3          & 32.6          & 66.7          & 60.7          & 29.8          & 18.2          & \textbf{27.7} & 40.5          & 5.9           & 11.4 \\
            SurroundOcc-style~\cite{surroundocc2023wei} & 37.5          & 30.5          & 68.4          & 56.9          & 28.2          & \textbf{18.9} & 27.2          & 26.8          & 7.0           & 10.9 \\
            \textbf{PanoBench (Ours)}                   & \textbf{58.9} & \textbf{52.0} & \textbf{82.8} & \textbf{82.3} & \textbf{54.8} & 18.4          & 27.6          & \textbf{45.3} & \textbf{59.1} & \textbf{46.0} \\
            \bottomrule
        \end{tabular}
    \end{table*}

    \begin{table}[!t]
        \centering
        \caption{Comparison of spatial continuity across occupancy benchmarks. Floating-voxel ratios are in permille (lower is better), and bold denotes the better result at the same voxel size.}
        \label{tab:spatial_continuity}
        \settablefont
        \setlength{\tabcolsep}{4pt}
        \begin{tabular}{llccc}
            \toprule
            Domain & Dataset                                    & Voxel Size (m) & $r_{\mathrm{fp}}@0.1$ & $r_{\mathrm{fp}}@0.2$ \\
            \midrule
            \multirow{4}{*}{CARLA}
                   & CarlaSC~\cite{motionsc2022wilson}          & 0.2            & 9.4                   & 15.1 \\
                   & \textbf{PanoBench-CarlaOcc}                & 0.2            & \textbf{2.5}          & \textbf{5.6} \\
                   & CarlaSC~\cite{motionsc2022wilson}          & 0.4            & 5.3                   & 11.8 \\
                   & \textbf{PanoBench-CarlaOcc}                & 0.4            & \textbf{0.3}          & \textbf{1.1} \\
            \midrule
            \multirow{4}{*}{nuScenes}
                   & OpenOccupancy~\cite{openoccupancy2023wang} & 0.2            & 23.2                  & 31.0 \\
                   & \textbf{PanoBench-nuScenes}                & 0.2            & \textbf{11.2}         & \textbf{17.4} \\
                   & Occ3D~\cite{occ3d2023tian}                 & 0.4            & 7.4                   & 16.9 \\
                   & \textbf{PanoBench-nuScenes}                & 0.4            & \textbf{1.2}          & \textbf{4.3} \\
            \midrule
            \multirow{2}{*}{KITTI-360}
                   & SSCBench-KITTI360~\cite{sscbench2024li}    & 0.2            & 16.3                  & 22.0 \\
                   & \textbf{PanoBench-KITTI360}                & 0.2            & \textbf{8.7}          & \textbf{11.8} \\
            \bottomrule
        \end{tabular}
    \end{table}

    \begin{table}[!t]
        \centering
        \caption{Comparison of temporal consistency across occupancy benchmarks. Bold denotes the better result at the same voxel size.}
        \label{tab:temporal_background_consistency}
        \settablefont
        \setlength{\tabcolsep}{4pt}
        \begin{tabular}{llccc}
            \toprule
            Domain & Dataset                                    & Voxel Size (m) & mIoU          & IoU \\
            \midrule
            \multirow{5}{*}{CARLA}
                   & \textbf{PanoBench-CarlaOcc}                & 0.1            & 81.9          & 83.3 \\
                   & CarlaSC~\cite{motionsc2022wilson}          & 0.2            & 70.3          & 83.5 \\
                   & \textbf{PanoBench-CarlaOcc}                & 0.2            & \textbf{82.8} & \textbf{89.1} \\
                   & CarlaSC~\cite{motionsc2022wilson}          & 0.4            & 71.5          & 86.8 \\
                   & \textbf{PanoBench-CarlaOcc}                & 0.4            & \textbf{83.6} & \textbf{92.3} \\
            \midrule
            \multirow{5}{*}{nuScenes}
                   & OpenOccupancy~\cite{openoccupancy2023wang} & 0.2            & 60.9          & \textbf{68.2} \\
                   & \textbf{PanoBench-nuScenes}                & 0.2            & \textbf{64.7} & 67.7 \\
                   & Occ3D~\cite{occ3d2023tian}                 & 0.4            & 59.9          & 64.7 \\
                   & \textbf{PanoBench-nuScenes}                & 0.4            & \textbf{67.8} & \textbf{71.9} \\
                   & SurroundOcc~\cite{surroundocc2023wei}      & 0.5            & 63.8          & 74.3 \\
            \midrule
            \multirow{2}{*}{KITTI-360}
                   & SSCBench-KITTI360~\cite{sscbench2024li}    & 0.2            & 55.0          & 60.1 \\
                   & \textbf{PanoBench-KITTI360}                & 0.2            & \textbf{61.8} & \textbf{66.0} \\
            \bottomrule
        \end{tabular}
    \end{table}

    \subsubsection{Spatial Continuity}
    \label{subsubsec:spatial_continuity}

    Spatial continuity measures how much of an occupancy annotation consists of small, fragmented components rather than coherent semantic structures.
    For each frame, occupied voxels of the same semantic class are grouped into connected components under 26-neighborhood connectivity.
    A component~$C$ is regarded as floating at volume threshold~$V$\,m$^{3}$ if $|C|\,r^{3}\le V$, where $r$ is the voxel size.
    The floating-voxel ratio~$r_{\mathrm{fp}}$ is then defined as follows:
    \begin{equation}
        r_{\mathrm{fp}}@V
        =
        \frac{\sum_{C}|C|\cdot\mathbb{I}[|C|\,r^{3}\le V]}{\sum_{C}|C|}.
    \end{equation}
    We report $r_{\mathrm{fp}}@0.1$ and $r_{\mathrm{fp}}@0.2$ for each occupancy dataset and compare them at matched voxel sizes.

    As shown in Table~\ref{tab:spatial_continuity}, PanoBench consistently reduces floating fragments at both volume thresholds across all data domains.
    On CARLA, PanoBench-CarlaOcc lowers $r_{\mathrm{fp}}@0.1$ and $r_{\mathrm{fp}}@0.2$ relative to CarlaSC~\cite{motionsc2022wilson} by 84.1\% and 76.7\%, respectively, averaged over the two voxel sizes.
    On real-world data, PanoBench reduces the floating-voxel ratios by 63.5\% on nuScenes and 46.5\% on KITTI-360 on average over both thresholds, compared with its counterparts~\cite{occ3d2023tian,openoccupancy2023wang,sscbench2024li} at the same voxel sizes.
    These results indicate that mesh-derived foregrounds and map-regularized backgrounds produce far fewer small semantic fragments than LiDAR aggregation pipelines.

    \subsubsection{Temporal Consistency}
    \label{subsec:temporal_consistency}

    Temporal consistency is quantified by comparing the ego-motion-aligned annotations of adjacent frames.
    For each consecutive keyframe pair, dynamic thing voxels are removed, the remaining background occupancy is warped from frame~$t$ into frame~$t{+}1$ with the sensor pose, and only voxels whose forward and backward projections stay inside both volumes are retained.
    For each dataset, 1,000 adjacent frame pairs are randomly sampled, and semantic mIoU and binary occupancy IoU are computed between the warped and original ground truth.

    As shown in Table~\ref{tab:temporal_background_consistency}, PanoBench is more temporally consistent than existing benchmarks in nearly all comparisons at matched voxel sizes.
    On CARLA, PanoBench-CarlaOcc improves mIoU over CarlaSC~\cite{motionsc2022wilson} by 12.5 and 12.1 points at 0.2\,m and 0.4\,m, respectively, together with IoU gains of 5.6 and 5.5 points.
    On nuScenes, PanoBench-nuScenes achieves the highest mIoU at both voxel sizes and improves IoU over Occ3D~\cite{occ3d2023tian} by 7.2 points at 0.4\,m, while its IoU at 0.2\,m is comparable to that of OpenOccupancy~\cite{openoccupancy2023wang}.
    SurroundOcc~\cite{surroundocc2023wei} attains the highest IoU of 74.3 among the nuScenes annotations, in line with the trend that coarser voxels yield higher consistency for every dataset evaluated at multiple voxel sizes.
    On KITTI-360, PanoBench-KITTI360 surpasses SSCBench-KITTI360~\cite{sscbench2024li} by 6.8 points in mIoU and 5.9 points in IoU at the same 0.2\,m voxel size.
    These results indicate that the scene-level background reconstruction of PanoBench remains stable under ego-motion, whereas aggregation-based annotations are more prone to surface flickering and inconsistent semantics.

\begin{figure*}[!t]
        \centering
        \includegraphics[width=0.99\textwidth]{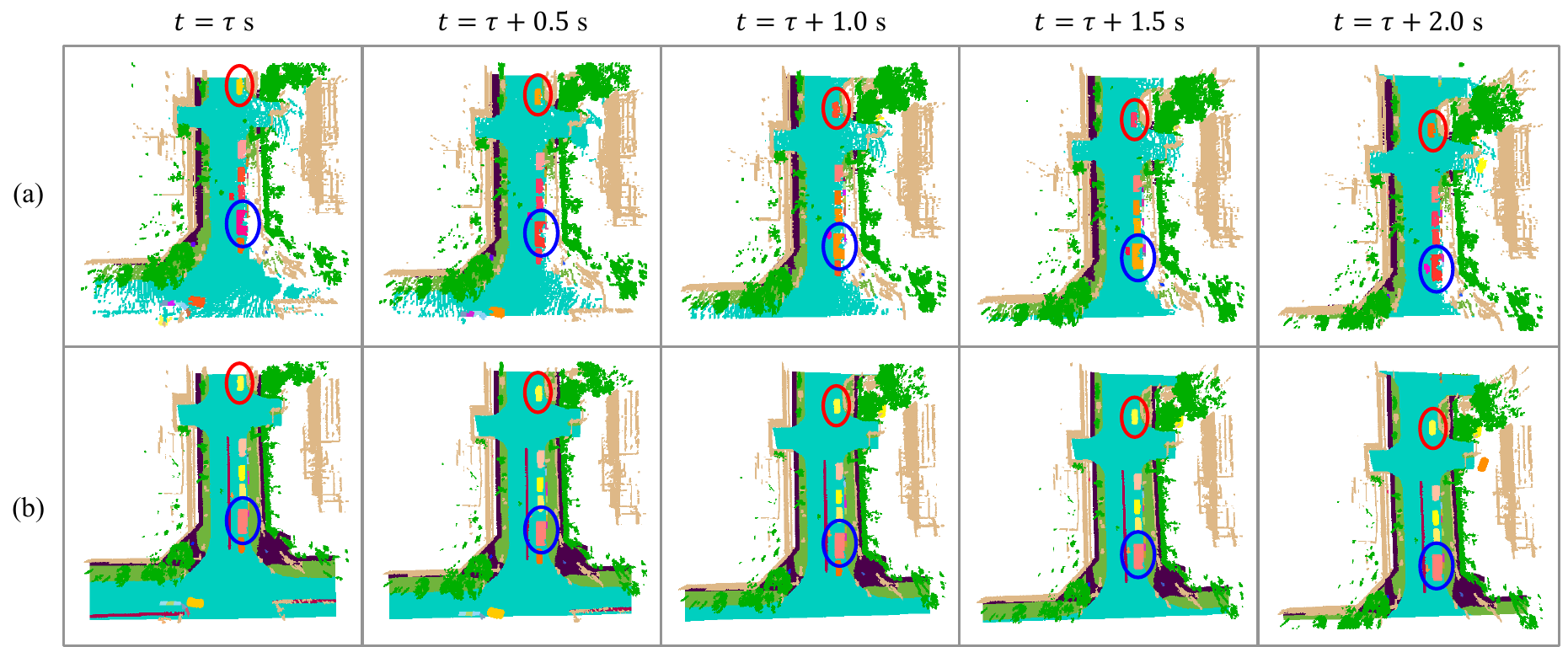}
        \caption{Comparison of instance temporal consistency on nuScenes: (a)~Occ3D-Panoptic~\cite{sparseocc2024liu} and (b)~PanoBench-nuScenes. Voxels with the same instance identity are rendered in the same color across frames, and red and blue circles highlight two representative instances.}
        \label{fig.compare_instance_temporal_consistency}
    \end{figure*}

    Fig.~\ref{fig.compare_instance_temporal_consistency} complements the background statistics with a qualitative view of foreground instance stability.
    Occ3D-Panoptic~\cite{sparseocc2024liu} exhibits noticeable temporal flickering, with object geometry and instance labels changing abruptly over time.
    In contrast, PanoBench preserves both foreground shape and instance identity throughout the observed trajectory.
    Overall, PanoBench provides temporally coherent annotations for both static backgrounds and dynamic foreground instances, enabling reliable supervision and evaluation for temporally aware panoptic occupancy prediction.

    \subsection{Ablation Study}
    \label{subsec:ablation_pipeline}

    The ablation study on the occupancy construction pipeline is conducted by successively enabling its background and foreground components under matched observations.
    Background variants are reconstructed on PanoBench-nuScenes and scored by observation fidelity on stuff pixels, following Sect.~\ref{subsubsec:observation_fidelity}.
    Foreground variants are reconstructed on PanoBench-CarlaOcc: each generated asset is voxelized in its annotated box and compared with the instance occupancy ground truth, so that occluded interiors are scored rather than visible silhouettes alone.

    \noindent\textbf{Background Reconstruction.}
    Starting from semantic LiDAR aggregation, we successively enable HD map-guided ground-layer lifting, topology-aware flat-region recovery, and non-planar refinement.
    As shown in Table~\ref{tab:ablation_background}, ground-layer lifting improves PixAcc from $68.5\%$ to $74.1\%$, recovering contiguous map-supported surfaces that sparse LiDAR returns cannot fill.
    Flat-region recovery adds another $4.9$ points, showing that map polygons still leave unlabeled holes and that topology-aware completion is required for continuous ground layers.
    Non-planar refinement contributes only $0.4$ points: aggregated LiDAR already covers most non-planar structures, so outlier filtering mainly removes floating artifacts rather than adding new geometry.
    Altogether, the full background pipeline improves PixAcc by $10.9$ points over semantic LiDAR aggregation alone.
    Because Occ3D~\cite{occ3d2023tian} and SurroundOcc~\cite{surroundocc2023wei} reconstruct occupancy primarily by aggregating and voxelizing LiDAR points, this gap indicates that map-guided ground completion provides a clear advantage over aggregation-based annotation protocols.

\begin{figure*}[!t]
        \centering
        \includegraphics[width=0.99\textwidth]{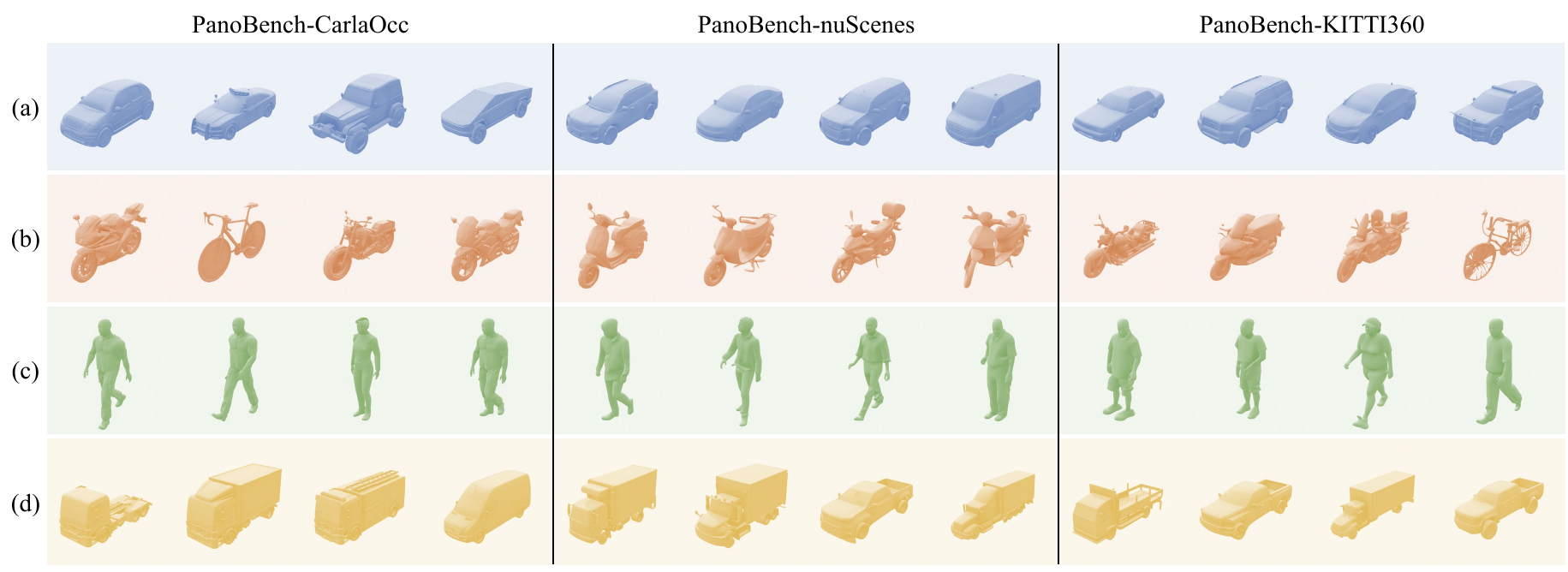}
        \caption{Representative prototype meshes from the three PanoBench subsets. Rows show (a)~cars, (b)~two-wheelers, (c)~pedestrians, and (d)~other vehicles. The examples illustrate the intra-class shape diversity retained by the prototype mesh pools across synthetic and real-world domains.}
        \label{fig.mesh_gallery}
    \end{figure*}

    \noindent\textbf{Foreground Mesh Asset Acquisition.}
    LiDAR aggregation inside each instance box~\cite{occ3d2023tian,surroundocc2023wei} is compared with image-conditioned mesh reconstruction under matched single-view and multi-view settings.
    Without view selection, every crop that passes the visibility test is used; the full setting retains only views that jointly satisfy the image-quality, mask-coverage, and multi-camera constraints.
    As shown in Table~\ref{tab:ablation_foreground}, LiDAR aggregation reaches only $12.4\%$ instance IoU, whereas even the weakest single-view generative model reaches $57.0\%$, confirming that aggregated LiDAR points cannot recover the complete geometry of foreground instances.
    Among single-view models, SAM~3D~\cite{sam3d2026chen} performs best, outperforming TRELLIS~\cite{trellis2024xiang} by $4.2$ points and TRELLIS-2~\cite{trellis22025xiang} by $0.7$ points, so a strong single-view model is a reliable fallback when multi-view observations are unavailable.
    The multi-view MV-SAM3D~\cite{li2026mv} further raises IoU to $65.1\%$ without view selection and $67.2\%$ with view selection, \ie, $3.9$ and $6.0$ points above the best single-view model.
    The $2.1$-point gain from view selection indicates that discarding occluded or low-quality observations is preferable to aggregating every available view.
    View-selected multi-view reconstruction is therefore used as the default foreground pipeline, with a fallback to SAM~3D~\cite{sam3d2026chen} when only one observation survives mining.
    Fig.~\ref{fig.mesh_gallery} shows representative prototype meshes from the three subsets, illustrating the intra-class shape diversity retained by the foreground pipeline.

    \begin{table}[!t]
        \centering
        \settablefont
        \caption{Ablation study on background reconstruction components on PanoBench-nuScenes. Observation fidelity is evaluated on stuff pixels.}
        \label{tab:ablation_background}
        \setlength{\tabcolsep}{3.0pt}
        \begin{tabular}{l|cccc}
            \toprule
            Background Component & \multicolumn{4}{c}{Configuration} \\
            \midrule
            Semantic LiDAR Aggregation & $\checkmark$ & $\checkmark$ & $\checkmark$ & $\checkmark$ \\
            Ground-Layer Lifting       &              & $\checkmark$ & $\checkmark$ & $\checkmark$ \\
            Flat-Region Recovery       &              &              & $\checkmark$ & $\checkmark$ \\
            Non-Planar Refinement      &              &              &              & $\checkmark$ \\
            \midrule
            $\mathrm{PixAcc}$ (\%)     & 68.5         & 74.1         & 79.0         & \textbf{79.4} \\
            \bottomrule
        \end{tabular}
    \end{table}

    \begin{table}[!t]
        \centering
        \settablefont
        \caption{Ablation study on foreground mesh asset acquisition on PanoBench-CarlaOcc, measured by instance IoU (\%).}
        \label{tab:ablation_foreground}
        \setlength{\tabcolsep}{3.0pt}
        \begin{tabular}{l|c|c|c}
            \toprule
            Method                                 & Input Views & View Selection & IoU \\
            \midrule
            LiDAR Aggregation~\cite{occ3d2023tian} & --          & --             & 12.4 \\
            \midrule
            SAM~3D~\cite{sam3d2026chen}            & Single-View & $\checkmark$   & 61.2 \\
            TRELLIS~\cite{trellis2024xiang}        & Single-View & $\checkmark$   & 57.0 \\
            TRELLIS-2~\cite{trellis22025xiang}     & Single-View & $\checkmark$   & 60.5 \\
            \midrule
            MV-SAM3D~\cite{li2026mv}               & Multi-View  & $\times$       & 65.1 \\
            MV-SAM3D~\cite{li2026mv}               & Multi-View  & $\checkmark$   & \textbf{67.2} \\
            \bottomrule
        \end{tabular}
    \end{table}

    \subsection{Occupancy Prediction Benchmarks}
    \label{subsec:semantic_panoptic_benchmark}
    \begin{table*}[t]
        \centering
        \settablefont
        \caption{Comparison of semantic occupancy prediction methods on PanoBench. Default voxel sizes are used, \ie, 0.4\,m for PanoBench-CarlaOcc and PanoBench-nuScenes and 0.2\,m for PanoBench-KITTI360.}
        \label{tab:main_semantic_results}
        \setlength{\tabcolsep}{2pt}
        \begin{tabular}{l|ccccc|ccccc|ccccc}
            \toprule
            \multirow{2}{*}{Method}
            & \multicolumn{5}{c|}{PanoBench-CarlaOcc}
            & \multicolumn{5}{c|}{PanoBench-nuScenes}
            & \multicolumn{5}{c}{PanoBench-KITTI360} \\
            \cmidrule(lr){2-6}\cmidrule(lr){7-11}\cmidrule(lr){12-16}
            & IoU           & mIoU          & RayIoU@1      & RayIoU@2      & RayIoU@4
            & IoU           & mIoU          & RayIoU@1      & RayIoU@2      & RayIoU@4
            & IoU           & mIoU          & RayIoU@1      & RayIoU@2      & RayIoU@4 \\
            \midrule
            SparseOcc~\cite{sparseocc2024liu}
            & 68.5          & 41.7          & 50.0          & 56.6          & 60.1
            & 55.0          & \textbf{33.0} & \textbf{33.7} & 41.7          & 46.5
            & 33.3          & 11.2          & \textbf{28.1} & \textbf{32.0} & \textbf{33.4} \\
            OPUS~\cite{opus2024wang}
            & 73.5          & 41.3          & 50.3          & 56.9          & 59.3
            & 50.1          & 29.9          & 32.6          & 41.4          & 46.2
            & 30.7          & 9.7           & 18.8          & 23.9          & 26.6 \\
            Panoptic-FlashOcc~\cite{flashocc2024yu}
            & 73.8          & 40.9          & 51.7          & 55.9          & 57.7
            & 47.7          & 22.6          & 25.2          & 34.9          & 41.1
            & 41.7          & 11.5          & 18.8          & 25.0          & 27.6 \\
            GaussianFormer~\cite{gaussianformer2024huang}
            & \textbf{82.2} & \textbf{46.2} & \textbf{56.4} & \textbf{61.0} & \textbf{62.8}
            & \textbf{66.5} & 32.6          & 32.8          & \textbf{41.8} & \textbf{46.7}
            & \textbf{65.3} & \textbf{15.6} & 21.1          & 25.5          & 27.1 \\
            \bottomrule
        \end{tabular}
    \end{table*}

    \begin{table*}[t]
        \centering
        \settablefont
        \caption{Comparison of panoptic occupancy prediction methods on PanoBench under the default voxel sizes in Table~\ref{tab:main_semantic_results}.}
        \label{tab:main_panoptic_results}
        \setlength{\tabcolsep}{2pt}
        \begin{tabular}{l|ccccc|ccccc|ccccc}
            \toprule
            \multirow{2}{*}{Method}
            & \multicolumn{5}{c|}{PanoBench-CarlaOcc}
            & \multicolumn{5}{c|}{PanoBench-nuScenes}
            & \multicolumn{5}{c}{PanoBench-KITTI360} \\
            \cmidrule(lr){2-6}\cmidrule(lr){7-11}\cmidrule(lr){12-16}
            & IoU           & mIoU          & RayPQ@1       & RayPQ@2       & RayPQ@4
            & IoU           & mIoU          & RayPQ@1       & RayPQ@2       & RayPQ@4
            & IoU           & mIoU          & RayPQ@1       & RayPQ@2       & RayPQ@4 \\
            \midrule
            SparseOcc~\cite{sparseocc2024liu}
            & 68.5          & \textbf{41.7} & \textbf{36.5} & \textbf{41.7} & \textbf{44.1}
            & \textbf{55.0} & \textbf{33.0} & \textbf{11.2} & \textbf{16.5} & \textbf{20.1}
            & 33.3          & 11.2          & \textbf{12.7} & \textbf{16.0} & \textbf{17.4} \\
            Panoptic-FlashOcc~\cite{flashocc2024yu}
            & \textbf{73.8} & 40.9          & 30.6          & 34.8          & 36.8
            & 47.7          & 22.6          & 3.2           & 7.0           & 10.8
            & \textbf{41.7} & \textbf{11.5} & 7.2           & 10.8          & 12.6 \\
            \bottomrule
        \end{tabular}
    \end{table*}

    \begin{table*}[t]
        \centering
        \settablefont
        \caption{Comparison of 4D occupancy forecasting methods on PanoBench under the default voxel sizes in Table~\ref{tab:main_semantic_results}.}
        \label{tab:results_occ_forecasting}
        \setlength{\tabcolsep}{3pt}
        \begin{tabular}{l|cccc|cccc|cccc|cccc|cccc|cccc}
            \toprule
            \multirow{3}{*}{Method}
            & \multicolumn{8}{c|}{PanoBench-CarlaOcc}
            & \multicolumn{8}{c|}{PanoBench-nuScenes}
            & \multicolumn{8}{c}{PanoBench-KITTI360} \\
            \cmidrule(lr){2-9}\cmidrule(lr){10-17}\cmidrule(lr){18-25}
            & \multicolumn{4}{c|}{Semantic mIoU} & \multicolumn{4}{c|}{Geometric IoU}
            & \multicolumn{4}{c|}{Semantic mIoU} & \multicolumn{4}{c|}{Geometric IoU}
            & \multicolumn{4}{c|}{Semantic mIoU} & \multicolumn{4}{c}{Geometric IoU} \\
            \cmidrule(lr){2-5}\cmidrule(lr){6-9}\cmidrule(lr){10-13}\cmidrule(lr){14-17}\cmidrule(lr){18-21}\cmidrule(lr){22-25}
            & 1\,s          & 2\,s          & 3\,s          & Avg.          & 1\,s          & 2\,s          & 3\,s          & Avg.
            & 1\,s          & 2\,s          & 3\,s          & Avg.          & 1\,s          & 2\,s          & 3\,s          & Avg.
            & 1\,s          & 2\,s          & 3\,s          & Avg.          & 1\,s          & 2\,s          & 3\,s          & Avg. \\
            \midrule
            SparseWorld~\cite{sparseworld2026dang}
            & 33.4          & 28.6          & 24.6          & 28.9          & 68.4          & 63.9          & 59.5          & 63.9
            & \textbf{25.3} & \textbf{23.1} & \textbf{21.0} & \textbf{23.1} & \textbf{49.2} & \textbf{47.6} & 45.7          & 47.5
            & 6.2           & 4.7           & 3.6           & 4.9           & 24.4          & 20.8          & 17.5          & 20.9 \\
            SparseWorld-TC~\cite{sparseworldtc2025du}
            & \textbf{34.5} & \textbf{29.8} & \textbf{25.7} & \textbf{30.0} & \textbf{69.3} & \textbf{64.8} & \textbf{60.4} & \textbf{64.8}
            & 24.9          & 22.7          & 20.8          & 22.8          & \textbf{49.2} & \textbf{47.6} & \textbf{45.9} & \textbf{47.6}
            & \textbf{7.7}  & \textbf{6.1}  & \textbf{4.8}  & \textbf{6.2}  & \textbf{28.6} & \textbf{24.8} & \textbf{21.1} & \textbf{24.8} \\
            \bottomrule
        \end{tabular}
    \end{table*}

    \subsubsection{Semantic Occupancy Prediction}
    Table~\ref{tab:main_semantic_results} reveals a clear divergence between volumetric and ray-based evaluation.
    GaussianFormer achieves the highest IoU on all three subsets and leads every metric on PanoBench-CarlaOcc, but its advantage does not carry over to RayIoU on real data: it is only on par with SparseOcc and OPUS on PanoBench-nuScenes, and on PanoBench-KITTI360 it is surpassed at every depth tolerance by SparseOcc, whose IoU is only about half as high.
    Rankings also shift across domains, \eg, Panoptic-FlashOcc ranks second in RayIoU@1 on PanoBench-CarlaOcc but last or tied for last on both real-world subsets.

    The two metric families reward different properties.
    Under the valid mask in Sect.~\ref{subsec:valid_mask}, IoU covers complete thing instances as well as observed stuff and free space, rewarding volumetric completeness, whereas RayIoU scores only the first occupied voxel along each camera ray, so a single voxel predicted in front of a visible surface invalidates a correct hit while barely affecting IoU.
    The dense Gaussian splatting of GaussianFormer fills occluded object parts but, under weak depth cues, tends to extend occupancy in front of surfaces, whereas SparseOcc keeps only a fixed number of the most confident voxels, which yields precise first hits but leaves much of the occluded volume empty on the finer grid of PanoBench-KITTI360.
    GaussianFormer is thus more complete and SparseOcc more precise, a trade-off absent only on the synthetic subset.

    \subsubsection{Panoptic Occupancy Prediction}
    Table~\ref{tab:main_panoptic_results} shows that instance quality does not follow volumetric or semantic accuracy.
    Although Panoptic-FlashOcc has a higher IoU and a nearly identical mIoU on PanoBench-CarlaOcc and PanoBench-KITTI360, SparseOcc achieves the higher RayPQ at every depth tolerance on all three subsets, most markedly on PanoBench-nuScenes, where its RayPQ@1 is over three times as high.
    For both methods, RayPQ@4 retains about two thirds or more of RayIoU@4 on PanoBench-CarlaOcc but about half or less on the real-world subsets, so grouping correctly classified surfaces into instances is the main difficulty in real scenes.

    The depth-tolerance dependence of RayPQ reveals why the two methods differ.
    Relaxing the tolerance from 1\,m to 4\,m raises the RayPQ of both methods by only about one fifth on PanoBench-CarlaOcc, but on PanoBench-nuScenes it more than triples that of Panoptic-FlashOcc while less than doubling that of SparseOcc, suggesting that many instances of Panoptic-FlashOcc are recognized but displaced in depth.
    This agrees with its design: thing voxels are grouped around instance centers predicted in the BEV plane after depth-based lifting, so depth errors displace entire instances, whereas SparseOcc predicts instance masks directly in 3D.

    \subsubsection{4D Occupancy Forecasting}
    As shown in Table~\ref{tab:results_occ_forecasting}, both methods degrade monotonically with the horizon on every subset, and semantics degrade faster than geometry; from 1\,s to 3\,s, SparseWorld-TC loses about a quarter of its mIoU but only an eighth of its IoU on PanoBench-CarlaOcc.
    IoU is dominated by static background, whose future occupancy follows largely from ego motion, whereas mIoU weights all classes equally and is sensitive to the future positions of small dynamic objects.
    The degradation is mildest on PanoBench-nuScenes, whose 80\,m surround-view volume remains largely within observed regions, and most severe on PanoBench-KITTI360, where more than a third of the 1\,s mIoU is lost by 3\,s, as a single forward-facing camera has seen much of the future volume only at long range or not at all.

    The two paradigms differ in how the future is decoded: SparseWorld~\cite{sparseworld2026dang} propagates the current sparse queries through state-conditioned regression, whereas SparseWorld-TC~\cite{sparseworldtc2025du} decodes every future frame directly from past image features with queries conditioned on the future ego trajectory.
    SparseWorld-TC is better at every horizon on PanoBench-CarlaOcc and PanoBench-KITTI360, slightly on the former and substantially on the latter, where its average mIoU and IoU are higher by roughly a quarter and a fifth, respectively, while the two are nearly indistinguishable on PanoBench-nuScenes.
    Trajectory-conditioned decoding thus helps most where past observations constrain the future least, yet does not prevent the degradation.

    \subsubsection{Discussion}
    These results lead to three insights.
    First, no single architecture dominates across metrics, because volumetric completeness and surface precision are distinct objectives.
    IoU alone can thus mis-rank methods, and combining dense coverage with sparse surface precision is a promising direction.
    Second, instance grouping, rather than semantic labeling, is the main bottleneck in real scenes, and its sensitivity to depth errors favors instance reasoning directly in 3D.
    Third, rankings do not reliably transfer from synthetic to real data, and limited camera coverage, as in PanoBench-KITTI360, hampers both current and future prediction.

    \subsection{Extended Experimental Results}
    \label{subsec:extended_results}

    \subsubsection{Multi-Resolution Analysis}
    \label{subsec:multi_resolution}
    PanoBench-CarlaOcc provides independently voxelized labels at 0.1, 0.2, and 0.4\,m, which makes it a suitable testbed for multi-resolution occupancy prediction.
    Each method is trained and evaluated separately at each native resolution.
    As shown in Table~\ref{tab:multi_resolution}, the task becomes steadily harder for a fixed architecture as the voxel size decreases.
    Averaged across the two methods~\cite{opus2024wang,flashocc2024yu}, reducing the voxel size from $0.4$\,m to $0.2$\,m decreases IoU and mIoU by 11.5 and 9.3 points, respectively.
    A further reduction to $0.1$\,m incurs substantially larger drops of 23.7 and 20.1 points.
    This sharper degradation arises because the image resolution and depth cues remain unchanged, while each halving of the voxel size multiplies the number of cells in the same physical volume by eight.
    Consequently, small depth or localization errors displace predictions across multiple voxels, and thin structures and semantic boundaries demand substantially greater geometric precision.
    Coarser grids absorb many of these errors through spatial quantization, whereas the $0.1$\,m grid penalizes them explicitly.

    \begin{table}[t]
        \centering
        \settablefont
        \caption{Multi-resolution occupancy prediction on PanoBench-CarlaOcc.}
        \label{tab:multi_resolution}
        \setlength{\tabcolsep}{3.5pt}
        \begin{tabular}{l|cccc}
            \toprule
            Method & Voxel Size (m) & IoU           & mIoU \\
            \midrule
            \multirow{3}{*}{OPUS~\cite{opus2024wang}}
                   & 0.1            & 38.5          & 11.7 \\
                   & 0.2            & 61.6          & 31.6 \\
                   & 0.4            & \textbf{73.5} & \textbf{41.3} \\
            \midrule
            \multirow{3}{*}{Panoptic-FlashOcc~\cite{flashocc2024yu}}
                   & 0.1            & 38.5          & 11.8 \\
                   & 0.2            & 62.8          & 32.1 \\
                   & 0.4            & \textbf{73.8} & \textbf{40.9} \\
            \bottomrule
        \end{tabular}
    \end{table}

    \subsubsection{Synthetic-to-Real Transfer Learning}
    \label{subsec:sim_to_real}
    Sim-to-real transfer is evaluated by comparing models pretrained on PanoBench-CarlaOcc and then fine-tuned on each target domain with models trained on the target domain from scratch.
    For source pretraining, we use the surround-view variant of PanoBench-CarlaOcc when transferring to PanoBench-nuScenes and its forward-view variant when transferring to PanoBench-KITTI360, matching the camera coverage of each target domain.
    Classes are mapped through shared semantic superclasses, and predictions are transformed to the target spatial range and voxel resolution before evaluation.
    Table~\ref{tab:cross_dataset_generalization} reports both pretraining settings under otherwise identical target-domain training and evaluation protocols.
    \begin{table}[t]
        \centering
        \settablefont
        \caption{Sim-to-real transfer results for panoptic occupancy prediction. RayIoU and RayPQ are averaged over depth tolerances of 1, 2, and 4\,m.}
        \label{tab:cross_dataset_generalization}
        \setlength{\tabcolsep}{2.0pt}
        \begin{tabular}{ll|c|ccc}
            \toprule
            Target                     & Method                                                   & Pretraining & mIoU & RayIoU & RayPQ \\
            \midrule
            \multirow{4}{*}{nuScenes}  & \multirow{2}{*}{SparseOcc~\cite{sparseocc2024liu}}       & --          & 33.0 & 40.6   & 15.9 \\
                                       &                                                          & CarlaOcc    & 34.2 & 42.3   & 17.4 \\
                                       & \multirow{2}{*}{Panoptic-FlashOcc~\cite{flashocc2024yu}} & --          & 22.6 & 33.7   & 7.0 \\
                                       &                                                          & CarlaOcc    & 26.4 & 35.6   & 9.6 \\
            \midrule
            \multirow{4}{*}{KITTI-360} & \multirow{2}{*}{SparseOcc~\cite{sparseocc2024liu}}       & --          & 11.2 & 31.2   & 15.4 \\
                                       &                                                          & CarlaOcc    & 11.8 & 31.9   & 15.2 \\
                                       & \multirow{2}{*}{Panoptic-FlashOcc~\cite{flashocc2024yu}} & --          & 11.5 & 23.8   & 10.2 \\
                                       &                                                          & CarlaOcc    & 12.1 & 25.0   & 10.9 \\
            \bottomrule
        \end{tabular}
    \end{table}

    CarlaOcc pretraining consistently improves mIoU and RayIoU for both architectures on both target domains.
    On PanoBench-nuScenes, SparseOcc gains 1.2 mIoU, 1.7 RayIoU, and 1.5 RayPQ points, while Panoptic-FlashOcc gains 3.8, 1.9, and 2.6 points, respectively.
    The transfer gains on PanoBench-KITTI360 are smaller: SparseOcc improves by 0.6 mIoU and 0.7 RayIoU but decreases by 0.2 RayPQ, whereas Panoptic-FlashOcc gains 0.6 mIoU, 1.2 RayIoU, and 0.7 RayPQ points.
    The larger gains on PanoBench-nuScenes are consistent with its surround-view coverage, which provides greater geometric overlap with the corresponding CarlaOcc pretraining setting.
    In contrast, PanoBench-KITTI360 offers less visual evidence and introduces larger differences in scene structure and label statistics, limiting the benefit of synthetic pretraining.
    The slight RayPQ decrease for SparseOcc further shows that improved semantic and geometric transfer does not necessarily preserve instance associations across domains.
    Overall, complete synthetic occupancy supervision provides a useful initialization, but its benefit depends on the alignment of camera coverage and instance distributions between the two domains.

    \subsubsection{Canonical Storage and I/O Efficiency}
    \label{subsec:canonical_storage_io}

    The practical benefit of the canonical sparse panoptic format introduced in Sect.~\ref{subsec:task_formulation} is evaluated by comparing the official annotation release with its converted sparse representation in terms of total storage footprint and per-frame I/O latency.
    \begin{table}[!t]
        \centering
        \settablefont
        \caption{Overall storage footprint and per-frame I/O latency of the official occupancy annotations versus our canonical sparse panoptic format. Latency is averaged over 1,000 randomly sampled frames per dataset on the same hardware.}
        \label{tab:canonical_storage_io}
        \setlength{\tabcolsep}{3.5pt}
        \begin{tabular}{lcccccc}
            \toprule
            \multirow{2}{*}{Dataset}
            & \multicolumn{3}{c}{Storage (GiB)}
            & \multicolumn{3}{c}{Latency (ms)} \\
            \cmidrule(lr){2-4}\cmidrule(lr){5-7}
            & Official & Ours         & $\downarrow$ (\%)
            & Official & Ours         & $\downarrow$ (\%) \\
            \midrule
            Occ3D-nuScenes~\cite{occ3d2023tian}
            & 2.6      & \textbf{1.3} & \textbf{48.7}
            & 11.1     & \textbf{1.6} & \textbf{85.8} \\
            Occ3D-Panoptic~\cite{sparseocc2024liu}
            & 3.2      & \textbf{1.3} & \textbf{58.1}
            & 20.6     & \textbf{1.5} & \textbf{92.8} \\
            OpenOccupancy~\cite{openoccupancy2023wang}
            & 132.7    & \textbf{5.5} & \textbf{95.9}
            & 40.2     & \textbf{6.8} & \textbf{83.2} \\
            SurroundOcc~\cite{surroundocc2023wei}
            & 40.3     & \textbf{1.2} & \textbf{97.1}
            & 13.5     & \textbf{1.9} & \textbf{85.8} \\
            SemanticKITTI~\cite{semantickitti2019behley}
            & 90.6     & \textbf{1.2} & \textbf{98.7}
            & 36.3     & \textbf{2.1} & \textbf{94.1} \\
            SSCBench-KITTI360~\cite{sscbench2024li}
            & 97.4     & \textbf{1.1} & \textbf{98.9}
            & 67.5     & \textbf{2.6} & \textbf{96.2} \\
            \bottomrule
        \end{tabular}
    \end{table}

    As shown in Table~\ref{tab:canonical_storage_io}, the canonical format consistently reduces both storage footprint and I/O latency.
    On dense releases such as OpenOccupancy~\cite{openoccupancy2023wang}, SurroundOcc~\cite{surroundocc2023wei}, SemanticKITTI~\cite{semantickitti2019behley}, and SSCBench-KITTI360~\cite{sscbench2024li}, storage drops by more than 95\% and per-frame latency by more than 83\%.
    Notably, Occ3D~\cite{occ3d2023tian} and its panoptic extension~\cite{sparseocc2024liu} already store annotations in compressed NPZ files, yet conversion to our canonical representation still reduces their storage footprints by approximately half, while decreasing loading latency by more than 85\%.
    The substantial latency reduction is largely attributable to the smaller serialized payload of the canonical format, which requires fewer bytes to be read and decoded for each frame.
    These results indicate that the sparse panoptic representation preserves the information required for training and evaluation while making large-scale experiments substantially more efficient.

    \section{Conclusion and Future Work}
    \label{sec:conclusion}
    This article presented PanoBench, a unified benchmark for camera-centric panoptic occupancy prediction across synthetic and real-world driving scenes.
    PanoBench-CarlaOcc contributes complete mesh-derived supervision, while PanoBench-nuScenes and PanoBench-KITTI360 reconstruct real scenes by combining semantic background geometry with complete instance-level foreground assets.
    A common sparse representation, temporally aware valid masks, and persistent instance identities make it possible to evaluate geometry, semantics, instance separation, temporal consistency, and domain transfer under one protocol.
    Quality analyses show that, at matched voxel sizes, PanoBench annotations are more observation-faithful, more complete, and less fragmented than those of existing benchmarks, and are more temporally consistent in nearly all comparisons.
    Benchmark results further reveal that no single architecture dominates across metrics and that instance-aware prediction on real-world data remains challenging.
    The benchmark is intended to shift occupancy evaluation from visible semantic surfaces toward complete, temporally supported 3D scene understanding.
    Future work will extend the benchmark to longer-horizon 4D panoptic forecasting and open-set categories, and study multimodal supervision while preserving camera-only inference.
    
    \normalem
    \bibliographystyle{IEEEtran}
    \bibliography{main}

\end{document}